\documentclass[10pt,twocolumn,letterpaper]{article}

\makeatletter
\def\input@path{{arXiv/}}
\makeatother

\usepackage[pagenumbers]{cvpr}

\usepackage{xspace}
\usepackage{algorithm}
\usepackage{algpseudocode}
\usepackage{xcolor}
\usepackage{amssymb}
\usepackage{booktabs}
\usepackage{multirow}
\usepackage{graphicx}
\usepackage{colortbl}
\usepackage{pgfplots}
\pgfplotsset{compat=1.18}

\definecolor{pcistage}{RGB}{43,145,170}
\definecolor{johteal}{RGB}{0,128,128}
\definecolor{kjhpink}{RGB}{210,70,130}
\definecolor{tabgreen}{RGB}{0,150,60}
\definecolor{tabred}{RGB}{200,30,30}
\definecolor{tabyellow}{RGB}{225,165,0}
\definecolor{tabfirst}{rgb}{1.0,0.6,0.6}
\definecolor{tabsecond}{rgb}{1.0,0.8,0.6}

\newcommand{\ours}{PerturbGS\xspace}

\newcommand{\cmark}{\textcolor{tabgreen}{\checkmark}}
\newcommand{\xmark}{\textcolor{tabred}{$\times$}}
\newcommand{\pmark}{\textcolor{tabyellow}{$\triangle$}}

\newcommand{\tabgroup}[1]{\multicolumn{6}{@{}l}{\textit{#1}}}

\newcommand{\fst}[1]{{\setlength{\fboxsep}{1pt}\colorbox{tabfirst}{\textbf{#1}}}}
\newcommand{\snd}[1]{{\setlength{\fboxsep}{1pt}\colorbox{tabsecond}{\underline{#1}}}}
\newcommand{\orc}[1]{\textcolor{gray}{#1}}
\newcommand{\nine}[1]{\rlap{$^{#1/9}$}}

\usepackage[resetlabels]{multibib}
\newcites{S}{Supplementary References}

\newcommand{\citesupp}[1]{%
    \begingroup
    \renewcommand{\citenumfont}[1]{S##1}%
    \citeS{#1}%
    \endgroup
}

\input{arXiv/cmlab_preamble}

\title{PerturbGS: Fast and Scalable Gaussian Splatting via Camera Perturbations}

\author{Seokhyun Youn$^{1}$ \quad Weeyoung Kwon$^{1}$ \quad Junho Kim$^{2}$ \quad Jihyong Oh$^{1,\dagger}$}

\cmlabAuthors{Seokhyun Youn$^{1}$, Weeyoung Kwon$^{1}$, Junho Kim$^{2}$, Jihyong Oh$^{1,\dagger}$}
\cmlabAffiliations{$^{1}$CMLab, Chung-Ang University \quad $^{2}$Cornell Tech \\ $^{\dagger}$Corresponding author}
\cmlabAuthorEmail{\{hisn16, weeyoungkwon, jihyongoh\}@cau.ac.kr \quad junho.kim@cornell.edu}
\cmlabProjectPage{https://cmlab-korea.github.io/PerturbGS}

\begin{document}

\twocolumn[{
    \renewcommand\twocolumn[1][]{#1}
    \maketitle
    \begin{center}
    \includegraphics[width=\textwidth]{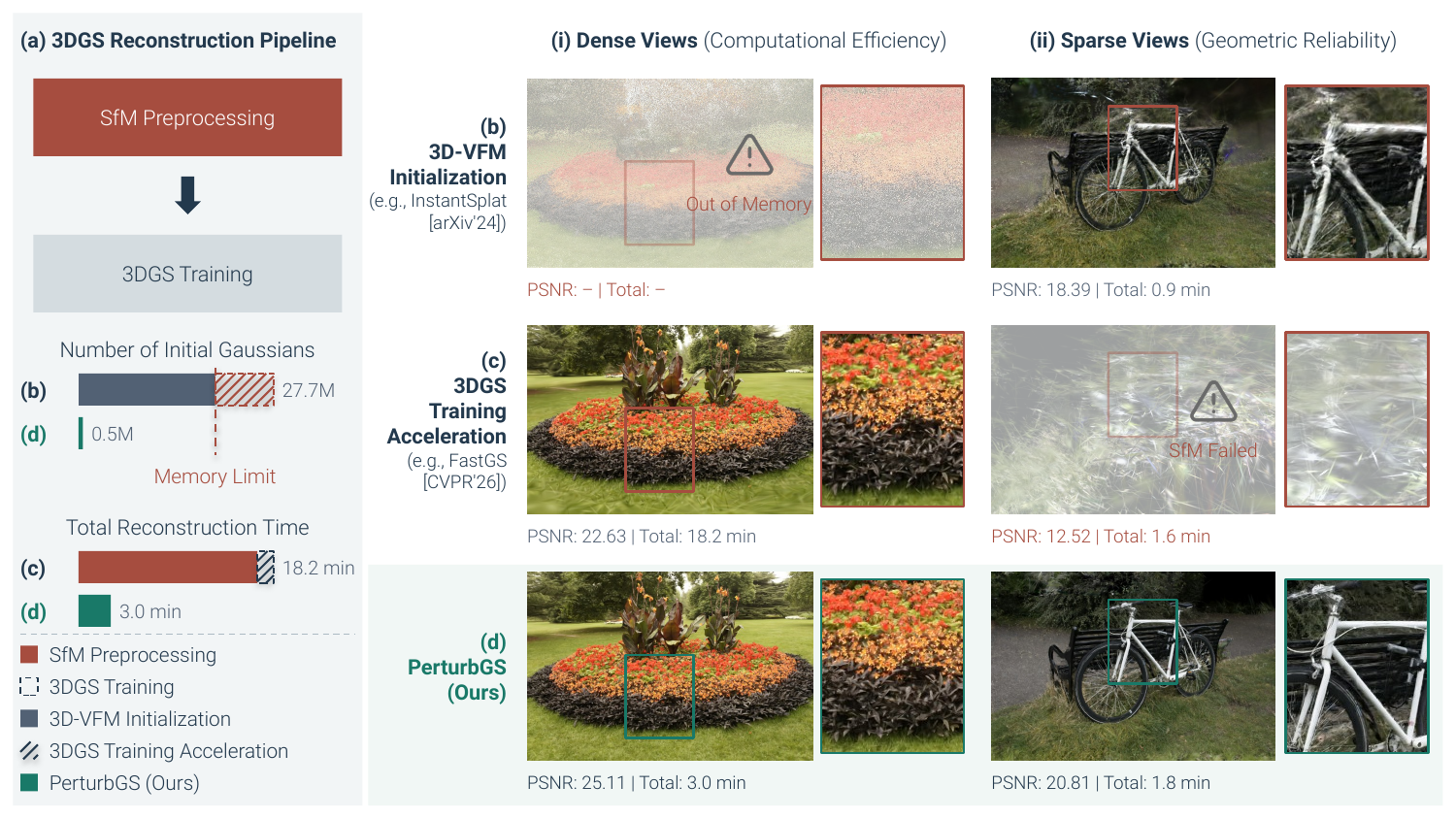}
    \captionof{figure}{%
        \textbf{Approaches for fast and scalable 3DGS reconstruction.}
(a) The 3DGS reconstruction pipeline consists of SfM preprocessing and 3DGS training.
The plots compare the number of initial Gaussians and the total reconstruction time.
Column (i) shows computational efficiency with dense views, and column (ii) shows geometric reliability with sparse views.
(b) 3D-VFM initialization yields too many Gaussians to fit in GPU memory in (i), and pose and depth errors cause artifacts in (ii).
(c) With 3DGS training acceleration, SfM preprocessing dominates the total reconstruction time in (i) and fails to register cameras in (ii).
(d) Our \ours{} completes 3DGS reconstruction with a more compact initialization than (b) and less time than (c) in (i), and renders a sharper novel view with reduced artifacts in (ii).
    }
    \label{fig:figure_page1}
\end{center}
\vspace{2mm}

}]

\begin{abstract}
Fast and scalable 3D Gaussian Splatting (3DGS) reconstruction from unposed input views requires both computational efficiency and geometric reliability.
With dense views, Structure-from-Motion (SfM) preprocessing dominates the total reconstruction time, and initializing Gaussians from the pixel-wise predictions of a 3D vision foundation model (3D-VFM) can exhaust GPU memory.
With sparse views, wide baselines can cause SfM registration to fail, and pose and depth errors in the predictions of the 3D-VFM can introduce artifacts in novel views.
We propose PerturbGS, a view-count-agnostic framework for fast and scalable 3DGS reconstruction from both dense and sparse views.
Our key insight is to use camera perturbations to expose geometric ambiguity before and during 3DGS training.
Before 3DGS training, our Camera-Perturbation-based Initialization (CPI) identifies depth-inconsistent point clouds through camera perturbations and removes them along with duplicate point clouds, yielding a compact initialization for 3DGS training.
During 3DGS training, our Camera-Perturbation-based Adaptive density control (CPA) prunes Gaussians with low rendering contribution across input and perturbed views.
Through camera perturbations, our CPA also groups Gaussians and densifies them group-wise, which limits the redundant growth of Gaussians.
Our PerturbGS reconstructs 3D scenes from dense views where previous methods exhaust GPU memory and produces sharper novel views with reduced artifacts from sparse views.
\end{abstract}

\section{Introduction}
\label{sec:intro}
3D Gaussian Splatting (3DGS)~\cite{kerbl20233d} reconstructs a 3D scene from multiple input images and enables real-time, high-quality rendering of novel views.
The 3DGS reconstruction pipeline (Fig.~\ref{fig:figure_page1}(a)) typically consists of two stages: Structure-from-Motion (SfM)~\cite{schonberger2016structure} preprocessing and 3DGS training, both of which contribute to the total reconstruction time.
To reduce the total reconstruction time, recent studies have pursued two directions, which we term \textit{3D vision foundation model (3D-VFM) initialization} and \textit{3DGS training acceleration}.
3D-VFM initialization (Fig.~\ref{fig:figure_page1}(b)) replaces SfM preprocessing with 3D-VFMs~\cite{wang2025vggt, wang2026pi, lin2025depth, wang2026vggt}, which jointly predict camera poses and scene geometry through feed-forward inference and provide a direct initialization for 3DGS training.
3DGS training acceleration (Fig.~\ref{fig:figure_page1}(c)) shortens 3DGS training through efficient density control, optimization, and rendering~\cite{ren2026fastgs, dhiman2024turbo, lyu2026faster}.

3D-VFM initialization and 3DGS training acceleration each improve only one stage of the 3DGS reconstruction pipeline.
This raises a fundamental question: ``\emph{Can improving either stage alone deliver fast, scalable, and high-quality 3DGS reconstruction regardless of the number of input views?}''
We examine 3D-VFM initialization and 3DGS training acceleration from two perspectives: \textit{computational efficiency} (Fig.~\ref{fig:figure_page1}(i)) and \textit{geometric reliability} (Fig.~\ref{fig:figure_page1}(ii)).

As shown in Fig.~\ref{fig:figure_page1}(i), computational efficiency degrades with dense views.
In 3D-VFM initialization, the pixel-wise predictions of the 3D-VFM from every input view accumulate into an excessive number of Gaussians, which causes out-of-memory (OOM) failures (Fig.~\ref{fig:figure_page1}(a),(b-i)).
Likewise, in 3DGS training acceleration, the runtime of SfM preprocessing grows with the number of input views and can dominate the total reconstruction time (Fig.~\ref{fig:figure_page1}(a),(c-i)).
As shown in Fig.~\ref{fig:figure_page1}(ii), geometric reliability degrades with sparse views.
In 3D-VFM initialization, wide baselines between input views provide insufficient multi-view cues for the 3D-VFM to predict accurate geometry~\cite{li2026wider}, and the inaccurate geometry produces artifacts (Fig.~\ref{fig:figure_page1}(b-ii)).
Likewise, in 3DGS training acceleration, wide baselines provide insufficient correspondences for SfM to recover camera poses and initial geometry reliably (Fig.~\ref{fig:figure_page1}(c-ii)).

Combining 3D-VFM initialization with 3DGS training acceleration eliminates SfM preprocessing, but the combination does not address the redundant and inaccurate predictions of the 3D-VFM (\textit{Suppl.}~\ref{sec:supp_vfm_acceleration}).
With dense views, duplicate point clouds initialize an excessive number of Gaussians, which causes OOM failures that degrade computational efficiency.
With sparse views, inaccurate geometry persists through 3DGS training and produces artifacts that degrade geometric reliability.

\begin{figure}[t]
  \centering
  \includegraphics[width=\linewidth]{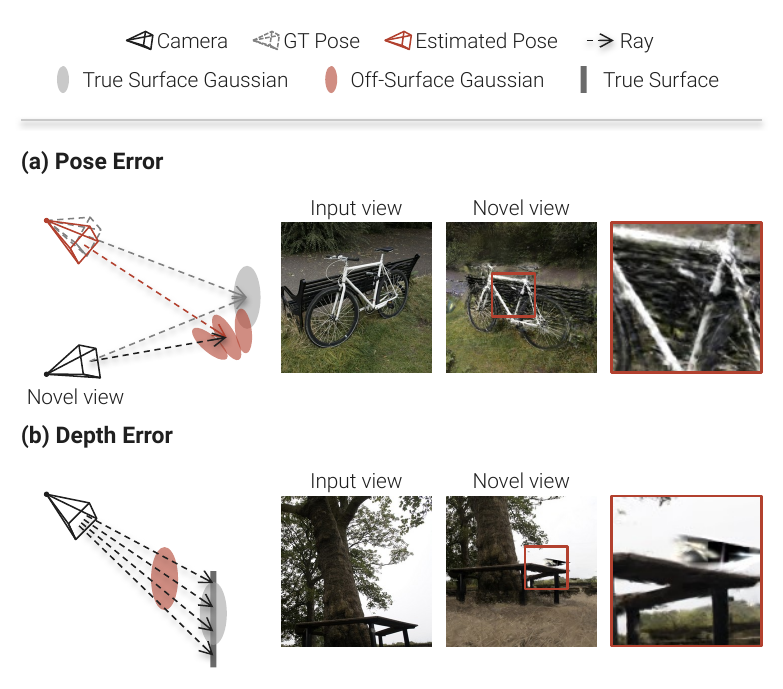}
  \caption{\textbf{Geometric ambiguity in 3D-VFM predictions.}
(a) \textbf{Pose error.} With an inaccurate pose, photometric supervision is applied along the wrong ray, causing 3DGS to fit the misalignment across views as view-dependent appearance.
(b) \textbf{Depth error.} An off-surface Gaussian can learn the appearance of the true surface in the input view but projects to a different pixel in the perturbed novel view.
Both errors can produce artifacts in novel views.}
  \label{fig:geometric_ambiguity}
\end{figure}

Our key insight is that the redundant and inaccurate predictions of the 3D-VFM stem from \textit{geometric ambiguity}.
Geometric ambiguity arises when the input views do not sufficiently constrain camera poses and depths.
We therefore perturb the camera of each input view to expose geometric ambiguity.
As shown in Fig.~\ref{fig:geometric_ambiguity}, pose and depth errors are hidden in the input view but appear in the perturbed novel view.

Based on this insight, we propose \ours{}, a view-count-agnostic framework that resolves geometric ambiguity before and during 3DGS training for fast, scalable, and high-quality 3DGS reconstruction from unposed input views.
To resolve geometric ambiguity before 3DGS training, we design our Camera-Perturbation-based Initialization (CPI), which coarsely refines the camera poses and depths predicted by the 3D-VFM and filters depth-inconsistent point clouds through camera perturbations, yielding a compact Gaussian initialization.
Since geometric ambiguity can remain during 3DGS training, we further design our Camera-Perturbation-based Adaptive density control (CPA), which prunes Gaussians with low rendering contribution across input and perturbed views and densifies Gaussians group-wise, limiting the redundant growth of Gaussians.

We summarize our contributions as follows:
\begin{itemize}
    \item We propose \ours{}, a view-count-agnostic framework that uses camera perturbations to resolve geometric ambiguity before and during 3DGS training for fast, scalable, and high-quality 3DGS reconstruction from unposed input views.
    \item We propose CPI, which coarsely refines the camera poses and depths predicted by the 3D-VFM and filters depth-inconsistent point clouds through camera perturbations.
    \item We propose CPA, which prunes Gaussians with low rendering contribution across input and perturbed views and densifies Gaussians group-wise.
    \item On Mip-NeRF~360 and Deep Blending, our \ours{} achieves fast, scalable, and high-quality 3DGS reconstruction from both dense and sparse views.
\end{itemize}

\section{Related Work}
\label{sec:related}

\noindent\textbf{3D-VFM initialization.}
3D-VFM initialization methods use predicted camera poses and per-pixel depths to back-project pixels from each input view into 3D and initialize 3DGS with the resulting point cloud.
When the input views do not sufficiently constrain pose and depth, pixels corresponding to the same surface point across views can be back-projected to different 3D locations, introducing redundancy into the point cloud~\cite{liu2025vggt, li2026wider}.
Existing methods filter predicted point clouds at the input viewpoints before training to reduce redundancy~\cite{fan2024instantsplat, liu2025vggt, ma2026ba} and jointly refine camera poses and Gaussians during training to improve alignment between the cameras and scene geometry~\cite{fan2024instantsplat, liu2025vggt, jena2025sparfels}.
As shown in Table~\ref{tab:positioning} (a), geometric ambiguity can persist when redundant or off-surface points project consistently across the input views.
To address this issue, our CPI refines camera poses when multi-view constraints are sufficient, uses camera perturbations to identify and remove depth-inconsistent point clouds, and merges repeated observations for Gaussian initialization.
\begin{figure*}[t]
  \centering
  \includegraphics[width=\textwidth,height=0.5\textheight,keepaspectratio]{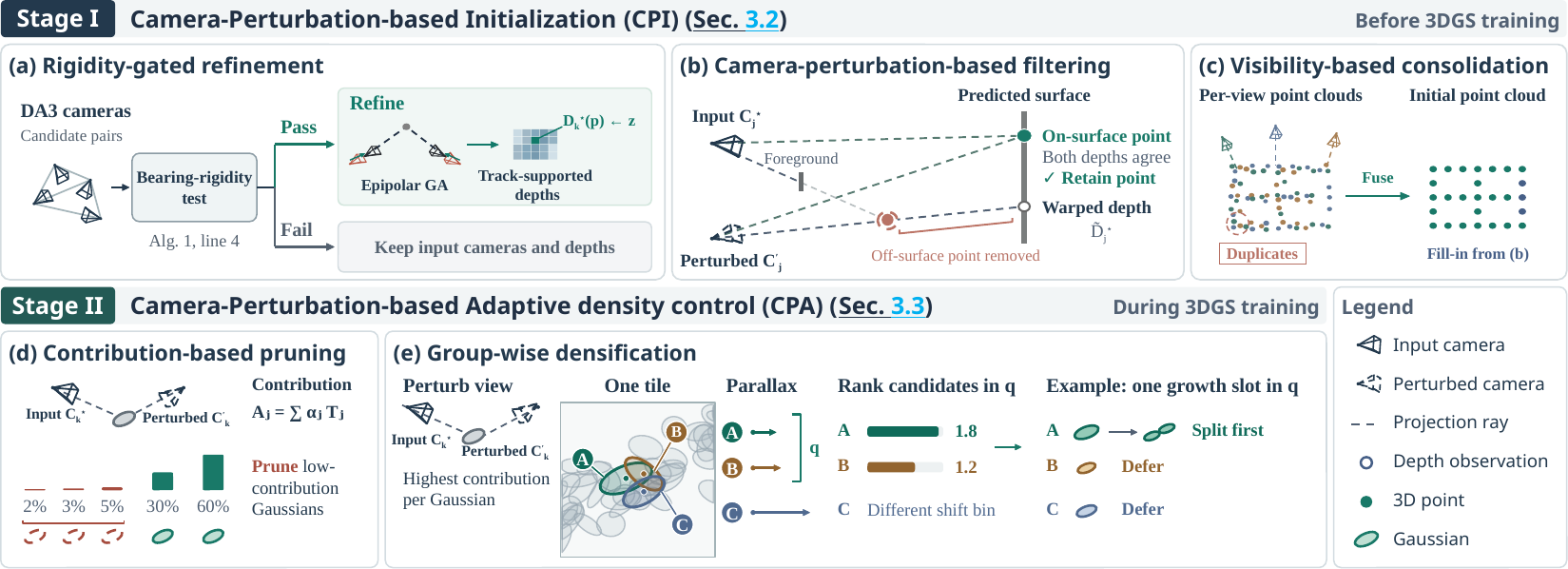}
  \caption{\textbf{Overview of \ours{}.}
Before training, Camera-Perturbation-based Initialization (CPI) constructs a compact initial point cloud.
(a) When the camera graph passes the bearing-rigidity test, CPI refines the camera poses predicted by DA3 and updates depth estimates supported by triangulated tracks.
(b) Camera perturbations expose off-surface points that are occluded in an input view. CPI filters depth-inconsistent points through cross-view depth voting.
(c) Visibility-based fusion merges repeated observations. Retained pixels outside the fusion groups provide fill-in points.
During training, Camera-Perturbation-based Adaptive density control (CPA) prunes Gaussians and limits redundant densification.
(d) CPA prunes Gaussians with low total rendering contribution across input and perturbed views.
(e) For each Gaussian, CPA selects the input view to which it contributes most and measures the displacement of its projected center under camera perturbation. Gaussians assigned to the same input view, image tile, and displacement bin share a densification budget. CPA admits split and clone candidates from SAD-GS in priority order until the next candidate would exceed the group budget.}
  \label{fig:overview}

\end{figure*}
\begin{table}[t]
  \centering
  \caption{\textbf{Conceptual comparison of approaches for fast and scalable 3DGS reconstruction.}
    Speed reflects total reconstruction time.
    3D-VFM initialization avoids SfM, but speed and memory efficiency vary by method.
    Training acceleration reduces optimization time but retains the cost of SfM.
    The feed-forward approach is fast and SfM-free but requires substantial memory.
    \ours{} addresses both pose and depth ambiguities while enabling fast, memory-efficient reconstruction without SfM.
    \cmark, \pmark, and \xmark\ denote full, partial, and no support, respectively.}
  \label{tab:positioning}
  \footnotesize
  \setlength{\tabcolsep}{4.5pt}
  \begin{tabular}{@{}lccccc@{}}
    \toprule
    & & \multicolumn{2}{c}{Efficiency} & \multicolumn{2}{c}{Geometric ambiguity} \\
    \cmidrule(lr){3-4} \cmidrule(lr){5-6}
    Method & SfM-free & Fast & Memory & \makebox[3.9em]{Pose} & \makebox[3.9em]{Depth} \\
    \midrule
    \tabgroup{(a) 3D-VFM initialization} \\
    InstantSplat~\cite{fan2024instantsplat} & \cmark & \cmark & \xmark & \pmark & \xmark \\
    VGGT-X~\cite{liu2025vggt}               & \cmark & \xmark & \cmark & \pmark & \pmark \\
    \midrule
    \tabgroup{(b) 3DGS Training acceleration} \\
    FastGS~\cite{ren2026fastgs}             & \xmark & \xmark & \cmark & \xmark & \xmark \\
    SAD-GS~\cite{lyu2026faster}             & \xmark & \xmark & \cmark & \xmark & \xmark \\
    \midrule
    \tabgroup{(c) Feed-forward 3DGS} \\
    AirSplat~\cite{bui2026airsplat}         & \cmark & \cmark & \xmark & \pmark & \pmark \\
    \midrule[\heavyrulewidth]
    \ours{} (\textbf{Ours})                              & \cmark & \cmark & \cmark & \cmark & \cmark \\ %
    \bottomrule
  \end{tabular}
  \vspace{-2mm}
\end{table}

\noindent\textbf{3DGS training acceleration.}
3DGS training acceleration methods start from camera poses and an initial point cloud, typically obtained by SfM~\cite{ren2026fastgs, mallick2024taming, chen2025dashgaussian, hollein20253dgs, dhiman2024turbo}.
Although these methods accelerate density control, optimization, and rendering, they still incur the cost of SfM preprocessing, which can dominate the total reconstruction time as the number of input views increases.
With sparse views, different Gaussian configurations can produce the same renderings at the input views~\cite{park2025dropgaussian, zhang2024cor}.
As shown in Tab.~\ref{tab:positioning}(b), existing 3DGS training acceleration methods leave this geometric ambiguity unresolved. Independently densifying Gaussians that explain the same image region can therefore lead to redundant growth.
In contrast, our CPA prunes Gaussians with low rendering contribution across input and perturbed views and densifies Gaussians group-wise, limiting the redundant growth of Gaussians.

\noindent\textbf{Feed-forward 3DGS.}
Feed-forward 3DGS methods use Gaussian predictors trained on large-scale data~\cite{lin2025depth, bui2026airsplat, ye2026yonosplat, kim2026f} to predict Gaussians from unposed input views in a single forward pass~\cite{jiang2025anysplat, park2025ecosplat}. Recent methods extend this approach to dense views~\cite{jiang2025anysplat, ye2026yonosplat, xu2026avsplat} or equip a 3D-VFM with a Gaussian head~\cite{lin2025depth, bui2026airsplat}.
However, as shown in Tab.~\ref{tab:positioning}(c), feed-forward prediction can leave geometric ambiguity unresolved, motivating methods such as AnySplat~\cite{jiang2025anysplat} and YoNoSplat~\cite{ye2026yonosplat} to refine the reconstruction through per-scene optimization initialized from the predicted Gaussians.
This additional optimization increases the total reconstruction time and can lead to OOM failures when the number of predicted Gaussians is large.
In contrast, our CPI reduces the number of initial Gaussians by filtering depth-inconsistent and duplicate point clouds, while our CPA limits the redundant growth of Gaussians by densifying Gaussians group-wise.

\section{Method}
\label{sec:method}

\subsection{Overview}
\label{sec:overview}
In 3D-VFM predictions, geometric ambiguity arises when different combinations of camera poses and scene geometry produce indistinguishable projections in the input views (Fig.~\ref{fig:geometric_ambiguity}).
With dense input views, repeated observations of the same surface can produce a redundant point cloud, leading to excessive Gaussian allocation and OOM failures during initialization. With sparse input views, inaccurate pose and depth estimates can remain undetected and degrade geometric reliability.
Our \ours{} therefore examines the geometry from perturbed cameras, which expose errors hidden at the input viewpoints, in two stages (Fig.~\ref{fig:overview}).
Stage I, Camera-Perturbation-based Initialization (CPI, Sec.~\ref{sec:pci}), refines the predicted cameras and removes off-surface and duplicate points before training.
Stage II, Camera-Perturbation-based Adaptive density control (CPA, Sec.~\ref{sec:CPA}), prunes and densifies Gaussians during training by their contribution to the input and perturbed views.
These two stages enable fast, geometrically reliable 3DGS reconstruction from both sparse and dense input views.

\subsection{Stage I (before training): CPI}
\label{sec:pci}
Back-projecting pixels from all input views can produce a point cloud in which pose and depth errors introduce off-surface points and repeated observations of the same surface introduce redundancy (Fig.~\ref{fig:geometric_ambiguity}). To address these issues, we design CPI to reduce geometric ambiguity in three steps before training, yielding a compact, geometrically reliable point cloud for 3DGS initialization from sparse or dense input views (Fig.~\ref{fig:cpi_pointcloud}, Alg.~\ref{alg:pci}).

\begin{algorithm}[t]
\caption{Camera-Perturbation-based Initialization (CPI).
Implementation details are provided in
\textit{Suppl.}~\ref{sec:supp_impl_pci}.}
\label{alg:pci}
\footnotesize
\begin{algorithmic}[1]

\Require Images $\mathcal I=\{I_k\}_{k=1}^{V}$;
    3D-VFM depths $\mathcal D$, confidence maps $\mathcal H$,
    and cameras $\mathcal C=\{(K_k,R_k,t_k)\}_{k=1}^{V}$
\Statex $V$: number of views; $K_k$: intrinsic parameters;
    $(R_k,t_k)$: extrinsic rotation and translation
\Ensure Points $S$, RGB colors $\mathbf c_S$,
    and cameras $\mathcal C^\star$

\Statex \textbf{(a) Rigidity-gated refinement}
\State $(\mathcal C^\star,\mathcal D^\star)
    \gets(\mathcal C,\mathcal D)$
    \label{line:pci_init}
\State $\mathcal E\gets
    \Call{CandidatePairs}{\mathcal C}$
\State $B\gets
    \Call{BearingMatrix}{\mathcal C,\mathcal E}$
    \Comment{\cite{zhao2015bearing}}
\If{$\operatorname{rank}(B)=3V-4$}
    \label{line:pci_gate}
    \State $\mathcal T\gets
        \Call{MatchViews}{\mathcal I,\mathcal C,\mathcal H,\mathcal E}$
    \State $\mathcal C^\star\gets
        \Call{EpipolarAlign}{\mathcal C,\mathcal H,\mathcal T}$
        \Comment{Global alignment (GA)}
    \State $\mathcal D^\star\gets
        \Call{TriangulateDepth}{\mathcal D,\mathcal C^\star,\mathcal T}$
\EndIf \label{line:pci_refine_end}

\Statex \textbf{(b) Camera-perturbation-based filtering}
\State $\mathcal C'\gets
    \Call{PerturbCameras}{\mathcal C^\star}$
    \label{line:pci_perturb}
\State $\widetilde{\mathcal D}^{\star}\gets
    \Call{ForwardWarp}{\mathcal D^\star,\mathcal C^\star,\mathcal C'}$
\State $\mathcal Q\gets
    \Call{DepthConsistencyVote}{
        \mathcal D^\star,\widetilde{\mathcal D}^{\star},
        \mathcal C^\star,\mathcal C'}$
    \label{line:pci_vote}

\Statex \textbf{(c) Visibility-based consolidation}
\State $(\mathcal F,\mathcal A)\gets
    \Call{VisibilityFusion}{\mathcal I,\mathcal D^\star,\mathcal C^\star}$
    \Comment{\cite{merrell2007real}}
    \label{line:pci_fusion}
\State $\mathcal P\gets
    \Call{Fill-In}{
        \mathcal F,\mathcal Q\setminus\mathcal A,
        \mathcal I,\mathcal D^\star,\mathcal C^\star}$
\State $(S,\mathbf c_S)\gets
    \Call{BudgetedSample}{\mathcal P}$
    \label{line:pci_sample}
\State \Return $S,\mathbf c_S,\mathcal C^\star$
    \label{line:pci_return}

\end{algorithmic}
\end{algorithm}

\noindent\textbf{(a) Rigidity-gated refinement.}
To address geometric ambiguity in the camera poses predicted by the 3D-VFM (Fig.~\ref{fig:geometric_ambiguity}(a)), CPI first determines whether the input views sufficiently constrain the relative camera positions.
{A pairwise bearing is the unit direction vector from one camera center to another. If these bearings leave relative camera spacing ambiguous, global alignment can preserve the bearings yet misalign scene geometry across views.
As shown in Stage~I of Fig.~\ref{fig:overview}(a), CPI applies a bearing-rigidity test~\cite{zhao2015bearing} to the graph of candidate camera pairs to decide whether to refine the poses (Alg.~\ref{alg:pci}, line~\ref{line:pci_gate}).
If the graph passes the test, the pairwise bearings determine the camera positions up to a global translation and scale. CPI then jointly refines the camera poses through epipolar global alignment~\cite{liu2025vggt} so that each matched point lies closer to the epipolar line predicted from its correspondence in the other view. Using the refined poses, CPI triangulates multi-view correspondences to update the depth estimates. Further details are provided in \textit{Suppl.}~\ref{sec:supp_impl_pci_refinement}.

\noindent\textbf{(b) Camera-perturbation-based filtering.}
To address geometric ambiguity in depth predictions, CPI checks each source pixel against depth estimates from other views. The source view alone cannot reveal a depth error, since back-projecting a pixel and reprojecting it into the same view recovers its predicted depth (Fig.~\ref{fig:geometric_ambiguity}(b)).
As described in Alg.~\ref{alg:pci}, CPI back-projects each source pixel using its 3D-VFM~\cite{lin2025depth} depth estimate after any selective updates from step~(a). It then projects the resulting 3D point into each other input camera \(C_j^\star\) and its perturbed counterpart \(C'_j\). At each projected location, CPI compares the point depth against the input depth map \(D_j^\star\) in \(C_j^\star\) or its forward-warped version \(\widetilde D_j^\star\) in \(C'_j\)~\cite{shade1998layered}.
As shown in Fig.~\ref{fig:overview}(b), camera perturbation can expose depth inconsistencies hidden by occlusion in an input view. Each camera pair casts a deletion vote if the reprojected depth is smaller than the map value in either view and a supporting vote if the depths agree in both views. CPI discards source pixels with more deletion votes than supporting votes and retains the rest in \(\mathcal Q\) (Alg.~\ref{alg:pci}, line~\ref{line:pci_vote}).
\begin{figure}[t]
    \centering
    \includegraphics[width=\linewidth]{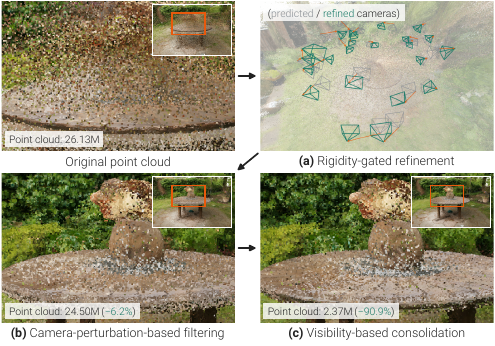}
\caption{\textbf{Camera-Perturbation-based Initialization (CPI).}
Starting from 3D-VFM predictions, CPI follows three steps in Alg.~\ref{alg:pci}:
(a) it refines camera poses when the rigidity test passes
(lines~\ref{line:pci_init}--\ref{line:pci_refine_end}),
(b) it filters depth-inconsistent points using camera perturbations
(lines~\ref{line:pci_perturb}--\ref{line:pci_vote}), and
(c) it fuses repeated observations into a compact point cloud
(lines~\ref{line:pci_fusion}--\ref{line:pci_return}).}
    \label{fig:cpi_pointcloud}
\end{figure}

\noindent\textbf{(c) Visibility-based consolidation.}
To prevent repeated observations of the same surface from initializing redundant Gaussians, CPI merges depth-consistent observations across overlapping views through visibility-based fusion~\cite{merrell2007real} (Fig.~\ref{fig:overview}(c), Alg.~\ref{alg:pci}, line~\ref{line:pci_fusion}). CPI then fills gaps in the fused point cloud by back-projecting pixels in \(\mathcal Q\setminus\mathcal A\), where \(\mathcal Q\) contains the pixels retained in (b) and \(\mathcal A\) contains those included in the fusion.
Finally, CPI uniformly subsamples the resulting point cloud to a fixed budget (Alg.~\ref{alg:pci}, line~\ref{line:pci_sample}). It passes the sampled 3D points \(S\), their RGB colors \(\mathbf c_S\), and the camera parameters \(\mathcal C^\star\) from step (a) to Stage II. We detail these steps in \textit{Suppl.}~\ref{sec:supp_pci_consolidation}.

\begin{table*}[t]
\centering
\caption{\textbf{Novel view synthesis across training view counts.}
We report PSNR$\uparrow$/SSIM$\uparrow$/LPIPS$\downarrow$ and the total time in minutes (Time$\downarrow$).
\textcolor{gray}{Gray} rows are oracles that start from the poses and point clouds of COLMAP ($^{\star}$) or GLOMAP ($^{\diamond}$), and they are excluded from the ranking.
The best results are in \fst{bold} and the second-best results are \snd{underlined}.
$^{n/N}$: number of scored scenes. $^{\dagger}$: mean including a wrong input reconstruction. OOM: out of memory.}
\label{tab:view_count}
\setlength{\tabcolsep}{3pt}
\def\metrichead{PSNR$\uparrow$/SSIM$\uparrow$/LPIPS$\downarrow$}
\resizebox{\textwidth}{!}{%
\def\venue#1{{\scriptsize [#1]}}
\def\two#1{\rlap{$^{#1/2}$}}
\def\viewhead{%
 & \multirow{2}{*}{Method}
 & \multicolumn{2}{c}{(i) 12 views} & \multicolumn{2}{c}{(ii) 24 views} & \multicolumn{2}{c}{(iii) 48 views}
 & \multicolumn{2}{c}{(iv) 96 views} & \multicolumn{2}{c}{(v) 192 views} \\
\cmidrule(lr){3-4} \cmidrule(lr){5-6} \cmidrule(lr){7-8} \cmidrule(lr){9-10} \cmidrule(lr){11-12}
 & & \metrichead & Time$\downarrow$ & \metrichead & Time$\downarrow$ & \metrichead & Time$\downarrow$
 & \metrichead & Time$\downarrow$ & \metrichead & Time$\downarrow$ \\}
\begin{tabular}{c l cc cc cc cc cc}
\toprule
\multicolumn{12}{c}{\textbf{Mip-NeRF~360}~\cite{barron2022mip}} \\
\midrule
\viewhead
\midrule
\orc{(a)} & \orc{FastGS$^{\star}$ \venue{CVPR'26}~\cite{ren2026fastgs}} &\orc{14.29$^{\dagger}$/0.292$^{\dagger}$/0.516$^{\dagger}$} & \orc{1.6} & \orc{20.15/0.572/0.305} & \orc{2.1} & \orc{24.35/0.746/0.204} & \orc{3.4} & \orc{27.56/0.843/0.151} & \orc{8.0} & \orc{28.48/0.863/0.134} & \orc{18.2} \\
\orc{(b)} & \orc{FastGS$^{\diamond}$ \venue{CVPR'26}~\cite{ren2026fastgs}} &\orc{14.07/0.263/0.543\nine{6}} & \orc{1.4} & \orc{16.92$^{\dagger}$/0.387$^{\dagger}$/0.431$^{\dagger}$\nine{8}} & \orc{1.7} & \orc{23.25/0.660/0.244} & \orc{2.4} & \orc{26.18/0.774/0.181} & \orc{4.3} & \orc{28.16/0.850/0.156} & \orc{8.4} \\
\orc{(c)} & \orc{SAD-GS$^{\star}$ \venue{SIGGRAPH'26}~\cite{lyu2026faster}} &\orc{14.47$^{\dagger}$/0.309$^{\dagger}$/0.512$^{\dagger}$} & \orc{0.8} & \orc{20.59/0.580/0.283} & \orc{1.4} & \orc{24.56/0.746/0.170} & \orc{2.6} & \orc{27.39/0.851/0.114} & \orc{7.1} & \orc{27.88/0.866/0.108} & \orc{17.1} \\
\orc{(d)} & \orc{SAD-GS$^{\diamond}$ \venue{SIGGRAPH'26}~\cite{lyu2026faster}} &\orc{14.37/0.273/0.550\nine{6}} & \orc{0.7} & \orc{16.35$^{\dagger}$/0.356$^{\dagger}$/0.460$^{\dagger}$} & \orc{1.0} & \orc{23.01/0.646/0.222} & \orc{1.7} & \orc{25.88/0.776/0.137} & \orc{3.6} & \orc{27.78/0.862/0.109} & \orc{7.5} \\
\cmidrule{1-12}
(e) & InstantSplat \venue{arXiv'24}~\cite{fan2024instantsplat} &17.34/0.407/0.433 & 0.9 & 18.03/0.452/0.460 & 2.4 & OOM & -- & OOM & -- & OOM & -- \\
(f) & VGGT-X \venue{arXiv'25}~\cite{liu2025vggt} &17.38/\snd{0.431}/\snd{0.414} & 7.2 & \snd{20.89}/\snd{0.583}/\snd{0.295} & 8.2 & \snd{23.99}/\snd{0.701}/\snd{0.212} & 9.5 & \snd{26.59}/\snd{0.802}/\snd{0.149} & 9.8 & \fst{27.60}/\snd{0.835}/\fst{0.131} & 10.0 \\
(g) & AirSplat \venue{ECCV'26}~\cite{bui2026airsplat} &\snd{18.17}/0.427/0.442 & $<$0.1 & 19.73/0.478/0.375 & $<$0.1 & 20.78/0.517/0.336 & $<$0.1 & 20.85/0.521/0.331 & $<$0.1 & 20.98/0.406/0.419\nine{2} & $<$0.1 \\
(h) & \textbf{\ours{} (Ours)} &\fst{18.38}/\fst{0.477}/\fst{0.375} & 1.8 & \fst{21.46}/\fst{0.620}/\fst{0.269} & 1.9 & \fst{24.32}/\fst{0.716}/\fst{0.200} & 2.1 & \fst{26.84}/\fst{0.817}/\fst{0.146} & 2.4 & \snd{27.51}/\fst{0.843}/\snd{0.136} & 3.0 \\
\midrule
\multicolumn{12}{c}{\textbf{Deep Blending}~\cite{hedman2018deep}} \\
\midrule
\viewhead
\midrule
\orc{(a)} & \orc{FastGS$^{\star}$ \venue{CVPR'26}~\cite{ren2026fastgs}} &\orc{11.01/0.434/0.645\two{1}} & \orc{1.3} & \orc{13.24/0.509/0.612} & \orc{1.4} & \orc{19.08$^{\dagger}$/0.674$^{\dagger}$/0.463$^{\dagger}$} & \orc{2.0} & \orc{24.67/0.802/0.332} & \orc{2.7} & \orc{27.78/0.853/0.281} & \orc{6.4} \\
\orc{(b)} & \orc{FastGS$^{\diamond}$ \venue{CVPR'26}~\cite{ren2026fastgs}} &\orc{--\two{0}} & \orc{--} & \orc{10.78$^{\dagger}$/0.391$^{\dagger}$/0.655$^{\dagger}$\two{1}} & \orc{1.4} & \orc{20.75/0.732/0.405\two{1}} & \orc{1.4} & \orc{25.34/0.828/0.313} & \orc{1.9} & \orc{29.19/0.891/0.264} & \orc{3.1} \\
\orc{(c)} & \orc{SAD-GS$^{\star}$ \venue{SIGGRAPH'26}~\cite{lyu2026faster}} &\orc{9.52$^{\dagger}$/0.261$^{\dagger}$/0.661$^{\dagger}$} & \orc{0.6} & \orc{12.80/0.484/0.627} & \orc{0.7} & \orc{18.77$^{\dagger}$/0.659$^{\dagger}$/0.466$^{\dagger}$} & \orc{1.2} & \orc{24.10/0.795/0.325} & \orc{2.0} & \orc{27.79/0.853/0.267} & \orc{5.7} \\
\orc{(d)} & \orc{SAD-GS$^{\diamond}$ \venue{SIGGRAPH'26}~\cite{lyu2026faster}} &\orc{--\two{0}} & \orc{--} & \orc{12.70$^{\dagger}$/0.577$^{\dagger}$/0.614$^{\dagger}$\two{1}} & \orc{0.5} & \orc{20.49/0.720/0.405\two{1}} & \orc{0.7} & \orc{25.28/0.822/0.309} & \orc{1.2} & \orc{28.90/0.892/0.247} & \orc{2.5} \\
\cmidrule{1-12}
(e) & InstantSplat \venue{arXiv'24}~\cite{fan2024instantsplat} &\snd{15.54}/\snd{0.680}/0.561 & 1.1 & 15.18/\fst{0.687}/0.576 & 2.5 & OOM & -- & OOM & -- & OOM & -- \\
(f) & VGGT-X \venue{arXiv'25}~\cite{liu2025vggt} &15.17/0.601/0.587\two{1} & 9.3 & \snd{18.08}/0.668/\snd{0.499} & 9.1 & \snd{19.18}/0.689/\snd{0.482} & 11.3 & 20.17/\fst{0.749}/0.475 & 14.0 & \snd{21.18}/\snd{0.760}/\snd{0.446} & 14.6 \\
(g) & AirSplat \venue{ECCV'26}~\cite{bui2026airsplat} &\fst{16.49}/\fst{0.689}/\fst{0.540} & $<$0.1 & 14.79/0.669/0.565 & $<$0.1 & 17.50/\snd{0.693}/0.511 & $<$0.1 & \snd{20.63}/0.736/\snd{0.440} & $<$0.1 & OOM & -- \\
(h) & \textbf{\ours{} (Ours)} &15.45/0.602/\snd{0.551} & 2.6 & \fst{18.39}/\snd{0.670}/\fst{0.478} & 2.8 & \fst{20.59}/\fst{0.716}/\fst{0.431} & 2.9 & \fst{21.02}/\snd{0.736}/\fst{0.425} & 3.3 & \fst{23.78}/\fst{0.788}/\fst{0.379} & 3.5 \\
\bottomrule
\end{tabular}}
\end{table*}

\subsection{Stage II (during training): CPA}
\label{sec:CPA}
After CPI, off-surface geometry can remain when depth estimates from multiple views agree at an incorrect 3D location. During training, independently densifying Gaussians that explain the same image region can introduce redundant Gaussians. CPA prunes Gaussians with low total rendering contribution across input and \textit{perturbed views}. Within each region, Gaussians with similar projected displacements under perturbation share a densification budget.

\noindent\textbf{(d) Contribution-based pruning.}
For view-count-agnostic pruning, CPA must assess Gaussian importance even when only a few input views are available. Contribution scores computed at the input viewpoints~\cite{fang2024mini} can undervalue a Gaussian that is occluded in those views but becomes visible from a nearby viewpoint. CPA therefore pairs each available input camera \(C_k^\star\) with a perturbed camera \(C'_k\) obtained by a small translation parallel to its image plane. This pairing tests visibility beyond the captured viewpoints for any number of input views. We sum the blending weights of Gaussian \(j\) over all camera pairs:
\begin{equation}
A_j =
\sum_{k=1}^{V}\sum_p
\left(
\alpha_j^{C_k^\star}(p)\,T_j^{C_k^\star}(p)
+
\alpha_j^{C'_k}(p)\,T_j^{C'_k}(p)
\right),
\label{eq:cpa_contribution}
\end{equation}
where \(\alpha_j^C(p)\) and \(T_j^C(p)\) denote the opacity and transmittance of Gaussian \(j\) at pixel \(p\) in camera \(C\). Before each densification step, CPA prunes Gaussians in ascending order of \(A_j\) until their cumulative contribution reaches a fraction \(\rho\) of the total.

\noindent\textbf{(e) Group-wise densification.}
CPA adopts the structure-tensor-guided split rule of SAD-GS~\cite{lyu2026faster} to reduce repeated splits during short training. This rule estimates in one step how many children a Gaussian needs to resolve local image detail. When overlapping Gaussians already resolve the same detail, applying the rule independently to each can produce redundant children.

To reduce this redundancy, CPA assigns each Gaussian to the input camera \(C_k^\star\) to which it contributes most. Within each camera, Gaussians that project to the same tile~\cite{kerbl20233d} in \(C_k^\star\) and exhibit similar image-plane displacements under perturbation to \(C'_k\) form a query \(q\) (Fig.~\ref{fig:overview}(e)). CPA applies the SAD-GS split rule to the representative Gaussian \(g_q\) of each query and uses the estimated child count as its shared densification budget \(N_q\). CPA processes split and clone candidates in descending order of priority and accepts an operation only if the resulting number of Gaussians in \(q\) does not exceed \(N_q\). Further details are provided in \textit{Suppl.}~\ref{sec:supp_cpa}.

\section{Experiments}
\label{sec:experiments}

\subsection{Experimental Setup}
\label{sec:experimental_setup}

\noindent\textbf{Datasets.}
We evaluate reconstruction across a wide depth range, from nearby objects to distant backgrounds, on the unbounded scenes of Mip-NeRF~360~\cite{barron2022mip} (\(8\times\) downsampling). Deep Blending~\cite{hedman2018deep} (original resolution) tests reconstruction under viewpoint-dependent occlusions in room-scale scenes captured from within.
For both datasets, we follow the view-sampling rule of DropGaussian~\cite{park2025dropgaussian} to construct splits ranging from 12 to 192 views.

\noindent\textbf{Evaluation protocol.}
To reflect practical settings where test views are unavailable, all methods use only the training views for SfM preprocessing and 3DGS training. For SfM-based methods, we follow SPARS3R~\cite{tang2025spars3r} and map the test camera poses from a separate SfM registration of all views with a similarity transform estimated on the training views. For SfM-free methods, we follow InstantSplat~\cite{fan2024instantsplat} and further refine each mapped test camera pose by minimizing the photometric \(L_1\) loss with the Gaussians fixed. We report PSNR, SSIM~\cite{wang2004image}, and LPIPS~\cite{zhang2018unreasonable} on the test views and the total reconstruction time. For a fair comparison, we measure all runtimes on a single NVIDIA RTX 5090 GPU.
\begin{figure}[t]
    \centering
    \includegraphics[width=\linewidth]{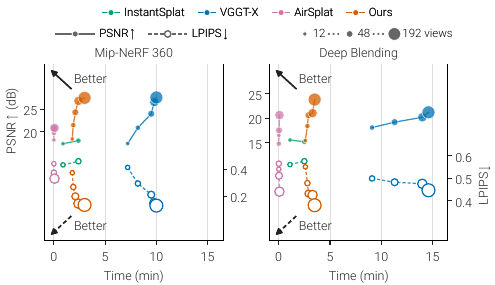}
    \caption{\textbf{Novel view quality against total reconstruction time.}
    The bubble area shows the number of input views (12 to 192), and a line joins the bubbles of one method as the number of input views increases.}
    \label{fig:time}
\end{figure}

\begin{figure*}[t]
    \centering
    \includegraphics[width=\textwidth]{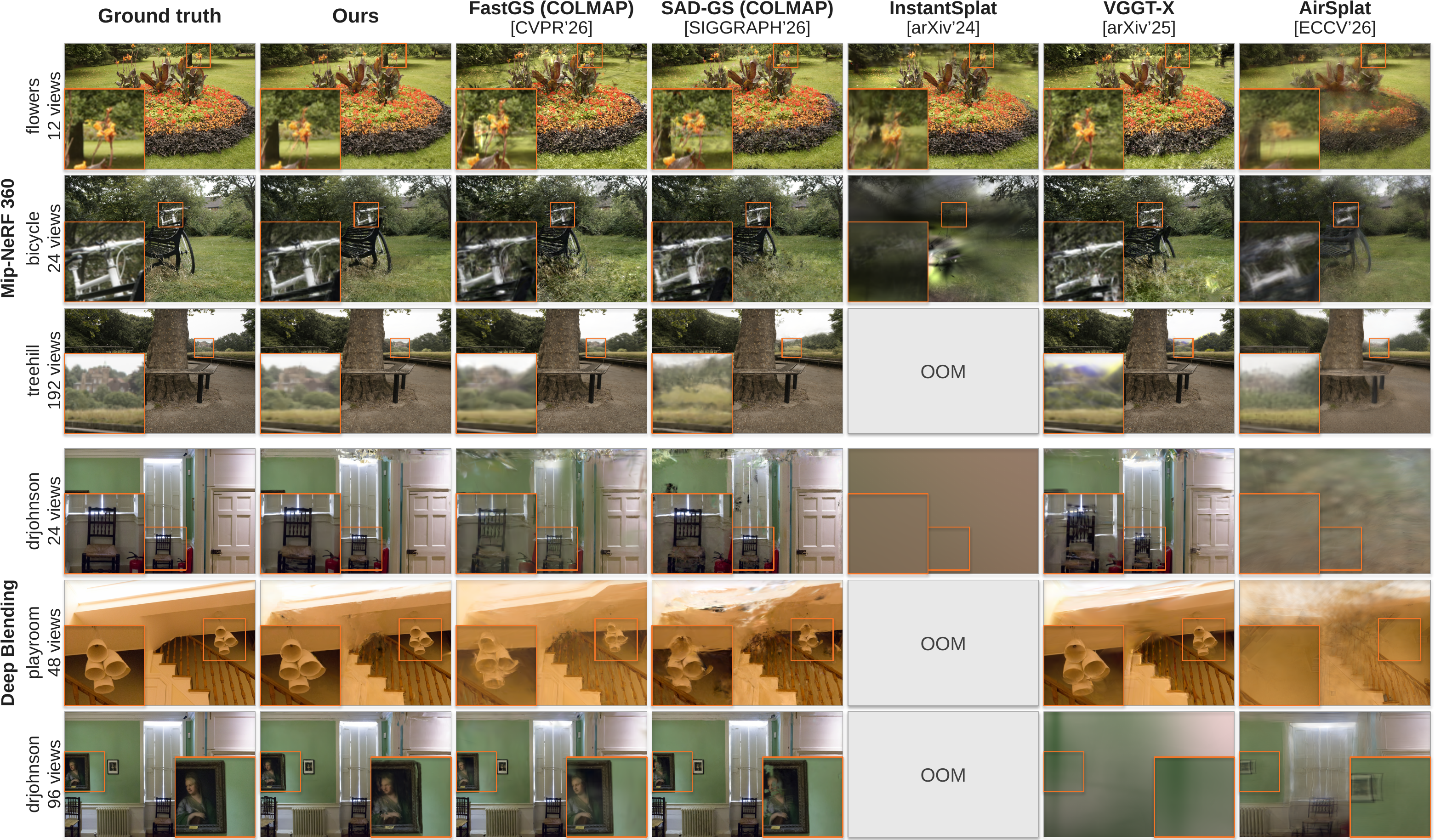}
    \caption{\textbf{Qualitative comparison on Mip-NeRF~360 and Deep Blending.}
    Insets show enlarged views of the marked regions. OOM denotes an out-of-memory error. See \textit{Suppl.}~\ref{sec:supp_demo} for additional qualitative comparisons.}
    \label{fig:qualitative}
\end{figure*}
\noindent\textbf{Comparison methods.}
We compare our \ours{} with three groups of methods across sparse and dense input views.
For \textit{3DGS training acceleration}, we use FastGS~\cite{ren2026fastgs} and SAD-GS~\cite{lyu2026faster} with COLMAP~\cite{schonberger2016structure,schonberger2016pixelwise} or GLOMAP~\cite{pan2024global} as SfM-based methods. For \textit{3D-VFM initialization}, we use InstantSplat~\cite{fan2024instantsplat} and VGGT-X~\cite{liu2025vggt} as SfM-free methods. For \textit{feed-forward 3DGS}, we use AirSplat~\cite{bui2026airsplat} as an SfM-free method without 3DGS training.

\subsection{Comparisons across View Counts}
\label{sec:view_count_comparison}

\noindent\textbf{Geometric reliability with sparse views.}
With 12 input views on Mip-NeRF~360, SfM-based acceleration methods using GLOMAP reconstruct only six of nine scenes (Tab.~2). In this setting, \ours{} achieves the best PSNR, SSIM, and LPIPS among all compared methods, with an average total reconstruction time of 1.8 minutes (Fig.~\ref{fig:time}). At 24 views, \ours{} leads on all three metrics on Mip-NeRF 360 and achieves the best PSNR and LPIPS on Deep Blending among the ranked methods. These results are consistent with CPI filtering out points with inconsistent depths across views before training. Compared with the feed-forward method AirSplat, \ours{} renders sharper flower contours in \textit{flowers} at 12 views and more distinct handlebars in \textit{bicycle} at 24 views (Fig.~\ref{fig:qualitative}), illustrating the detail achievable with a short 3DGS optimization.

\noindent\textbf{Computational efficiency with dense views.}
As the number of input views increases from 48 to 192 on Mip-NeRF~360, average reconstruction time rises from 3.4 to 18.2 minutes for FastGS with COLMAP and from 2.1 to 3.0 minutes for \ours{} (Tab.~\ref{tab:view_count}). CPI bounds the initial Gaussian count, and CPA limits redundant growth during training, allowing \ours{} to reconstruct every scene in both datasets without exhausting GPU memory. On Deep Blending, \ours{} achieves the best PSNR, SSIM, and LPIPS at 192 views in just 3.5 minutes on average. Fig.~\ref{fig:qualitative} shows novel views rendered from reconstructions with 48, 96, and 192 input views.

\subsection{Ablation Studies}
\label{sec:ablations}

\noindent\textbf{(a) Rigidity-gated refinement.}
Row (a) of Tab.~\ref{tab:ablation} shows that removing pose refinement leaves PSNR unchanged at 24 views but lowers it by 0.60~dB at 192 views.
To explain this difference, Tab.~\ref{tab:analysis_gate} compares three pose refinement strategies. Our method refines poses only in scenes that pass the bearing rigidity test. All Pass refines poses in every scene, while All Fail skips refinement in every scene and corresponds to row (a) of Tab.~\ref{tab:ablation}.
With sparse views, a camera can move while preserving the directions between camera pairs, as illustrated on the left. Global alignment can then degrade the predicted poses.
At 12 and 24 views, all nine scenes fail the rigidity test. Our method therefore retains the predicted poses in all nine scenes and achieves higher PSNR than All Pass in both settings (Tab.~\ref{tab:analysis_gate}(i)).
At 48 views, only four scenes pass, and refining these scenes yields higher PSNR than either alternative.
At 96 and 192 views, the camera directions constrain the positions in six and eight scenes, respectively. Refining those scenes improves PSNR over All Fail by 0.40 and 0.60~dB, respectively, and brings our gate within 0.13~dB of All Pass.

\begin{table}[t]
    \centering
    \caption{\textbf{Component ablation on Mip-NeRF 360.}
    Each row removes one component of our CPI, (a)--(c), or of our CPA, (d)--(e), from the full model.
    {\#G is the Gaussian count in millions (M), with its change relative to the full model in parentheses.}}
    \label{tab:ablation}

    \fontsize{6.5}{7.5}\selectfont
    \setlength{\tabcolsep}{1pt}
    \renewcommand{\arraystretch}{0.9}
    \def\dg#1{\,{\fontsize{5}{6}\selectfont\textcolor{gray}{(#1)}}}
    \resizebox{\columnwidth}{!}{%
    \begin{tabular}{@{}cl*{4}{c}@{}}
        \toprule
        & & \multicolumn{2}{c}{(i) 24 views}
        & \multicolumn{2}{c}{(ii) 192 views} \\
        \cmidrule(lr){3-4}
        \cmidrule(lr){5-6}
        & Variant
        & PSNR/SSIM/LPIPS & \#G (M)
        & PSNR/SSIM/LPIPS & \#G (M)\\
        \midrule
        & Baseline (w/o CPI and CPA)
        & 21.08/0.602/0.283 & 2.59\dg{$+$156\%} & 26.89/0.821/0.136 & 3.45\dg{$+$185\%} \\
        \midrule
        (a) & w/o rigidity-gated refinement
        & 21.46/0.620/0.269 & 1.01\dg{0\%} & 26.91/0.826/0.146 & 1.18\dg{$-$2\%} \\
        (b+c) & w/o filtering and consolidation
        & 21.24/0.605/0.282 & 0.96\dg{$-$5\%} & 27.43/0.843/0.137 & 1.21\dg{0\%} \\
        \midrule
        (d) & w/o contribution-based pruning
        & 21.32/0.615/0.270 & 1.48\dg{$+$47\%} & 27.61/0.841/0.129 & 2.44\dg{$+$102\%} \\
        (e) & w/o group-wise densification
        & 21.39/0.621/0.268 & 1.40\dg{$+$39\%} & 27.66/0.848/0.126 & 1.61\dg{$+$33\%} \\
        \midrule
        \rowcolor{gray!15}
        & \textbf{\ours{} (Full)}
        & 21.46/0.620/0.269 & 1.01 & 27.51/0.843/0.136 & 1.21 \\
        \bottomrule
    \end{tabular}%
    }
\end{table}

\begin{table}[t]
    \centering
    \caption{\textbf{Behavior of the rigidity gate on Mip-NeRF 360.}
    Left: the gate runs the refinement when the pairwise bearings (blue) determine every camera position and skips it when a camera (red) can move without changing any bearing.
    Values are PSNR, with the number of scenes on which global alignment runs in parentheses. $^{n/N}$: number of scored scenes. The best results are in \textbf{bold} and the second-best results are \underline{underlined}.}
    \label{tab:analysis_gate}
    \begin{minipage}[c]{0.33\columnwidth}
        \centering
        \includegraphics[width=\linewidth]{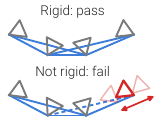}
    \end{minipage}\hfill
    \begin{minipage}[c]{0.64\columnwidth}
        \centering
        \footnotesize
        \setlength{\tabcolsep}{3pt}
        \resizebox{\linewidth}{!}{%
        \begin{tabular}{@{}clccc@{}}
            \toprule
            & Views & (i) \ours{} & (ii) All pass & (iii) All fail \\
            \midrule
            (a) & 12  & \textbf{18.38} (0/9) & 18.24$^{8/9}$ (9/9) & \underline{18.38} (0/9) \\
            (b) & 24  & \textbf{21.46} (0/9) & 20.88 (9/9) & \underline{21.46} (0/9) \\
            (c) & 48  & \textbf{24.32} (4/9) & 24.13 (9/9) & \underline{24.19} (0/9) \\
            (d) & 96  & \underline{26.84} (6/9) & \textbf{26.97} (9/9) & 26.44 (0/9) \\
            (e) & 192 & \underline{27.51} (8/9) & \textbf{27.56} (9/9) & 26.91 (0/9) \\
            \bottomrule
        \end{tabular}}
    \end{minipage}
\end{table}

\noindent\textbf{(b+c) Filtering and consolidation.}
As shown in Fig.~\ref{fig:cpi_pointcloud}, our filtering removes points with inconsistent depths across views, and our consolidation merges repeated observations of the same surface.
Together, these steps produce a more compact initial point cloud while preserving surface structure supported by multiple views.
This matters particularly with sparse views, where training provides few geometric constraints to correct errors in the initial geometry.
Row (b+c) of Tab.~\ref{tab:ablation} shows that removing filtering and consolidation degrades novel view quality across all three metrics at 24 views but has little effect at 192 views.
The larger improvement at 24 views suggests that our filtering and consolidation are particularly effective when reconstruction relies more heavily on the geometry established during initialization.

\noindent\textbf{(d) Contribution-based pruning.}
At 24 views, removing our pruning increases the Gaussian count by 47\% and worsens PSNR, SSIM, and LPIPS (Tab.~\ref{tab:ablation}(d)).
With sparse views, Gaussians at different depths can overlap in a training image. A small camera translation shifts their projections by different amounts, allowing our pruning to assess their contributions from a nearby viewpoint. The sparse-view results suggest that removing Gaussians with low total contribution across training and perturbed views improves novel view quality.
At 192 views, removing our pruning more than doubles the Gaussian count (+102\%) for only a small PSNR gain. Fig.~\ref{fig:ablation_cpa}(a) shows a region where our pruning preserves comparable visual quality with half as many Gaussians.
\textit{Suppl.}~\ref{sec:supp_analysis} compares our approach with pruning based on training views alone.
\begin{figure}[t]
    \centering
    \includegraphics[width=\linewidth]{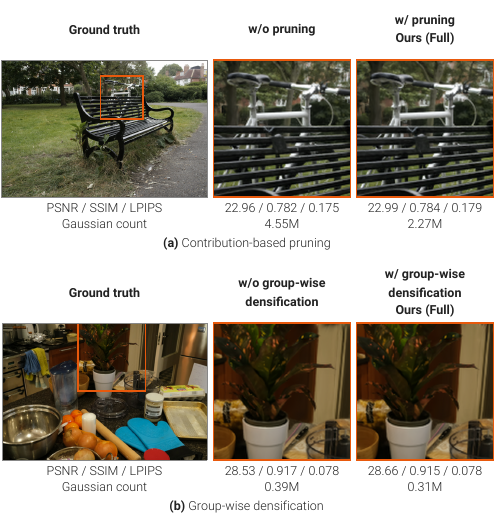}
    \caption{\textbf{Effect of CPA with 192 input views.}
For the region marked in each ground-truth image, we compare the rendering obtained by disabling (a) contribution-based pruning or (b) group-wise densification with that of the full model.
PSNR, SSIM, and LPIPS are computed over the full test image. Gaussian counts are reported per scene.}
    \label{fig:ablation_cpa}
\end{figure}

\noindent\textbf{(e) Group-wise densification.}
Removing the shared budget increases the Gaussian count by 39\% at 24 views and 33\% at 192 views (Tab.~\ref{tab:ablation}, row (e)). The additional Gaussians lower PSNR with sparse views and yield only a small gain with dense views. Fig.~\ref{fig:ablation_cpa}(b) shows a region where our budget produces a more accurate rendering with fewer Gaussians.
The shared budget limits how many children a group can add to the same image region and gives priority to the highest-ranked densification candidates. This curbs repeated splits or clones of Gaussians that remain difficult to distinguish under camera perturbation, while still allowing the group to refine local detail.

\section{Conclusion}
\label{sec:Conclusion}
We present \ours{}, a view-count-agnostic framework for fast 3DGS reconstruction from unposed images. Before training, CPI refines camera poses when the input views sufficiently constrain them, uses camera perturbations to filter depth-inconsistent points, and merges repeated observations to form a compact initial point cloud. During training, CPA prunes Gaussians with low rendering contribution across input and perturbed views and limits redundant allocation by assigning a shared densification budget to Gaussians with similar projected displacements. Experiments on Mip-NeRF~360 and Deep Blending show that \ours{} achieves high novel view quality with sparse inputs and maintains short reconstruction times as the number of views increases.

{
    \small
    \bibliographystyle{arXiv/ieeenat_fullname}
    \bibliography{arXiv/main}

\begin{thebibliography}{1}
\providecommand{\natexlab}[1]{#1}
\providecommand{\url}[1]{\texttt{#1}}
\expandafter\ifx\csname urlstyle\endcsname\relax
  \providecommand{\doi}[1]{doi: #1}\else
  \providecommand{\doi}{doi: \begingroup \urlstyle{rm}\Url}\fi

\bibitem[Potje et~al.(2024)Potje, Cadar, Araujo, Martins, and
  Nascimento]{potje2024xfeat}
Guilherme Potje, Felipe Cadar, Andr{\'e} Araujo, Renato Martins, and Erickson~R
  Nascimento.
\newblock Xfeat: Accelerated features for lightweight image matching.
\newblock In \emph{CVPR}, 2024.

\end{thebibliography}


\begin{thebibliography}{36}
\providecommand{\natexlab}[1]{#1}
\providecommand{\url}[1]{\texttt{#1}}
\expandafter\ifx\csname urlstyle\endcsname\relax
  \providecommand{\doi}[1]{doi: #1}\else
  \providecommand{\doi}{doi: \begingroup \urlstyle{rm}\Url}\fi

\bibitem[Barron et~al.(2022)Barron, Mildenhall, Verbin, Srinivasan, and
  Hedman]{barron2022mip}
Jonathan~T Barron, Ben Mildenhall, Dor Verbin, Pratul~P Srinivasan, and Peter
  Hedman.
\newblock Mip-nerf 360: Unbounded anti-aliased neural radiance fields.
\newblock In \emph{CVPR}, 2022.

\bibitem[Bui et~al.(2026)Bui, Moon, and Kim]{bui2026airsplat}
Minh-Quan~Viet Bui, Jaeho Moon, and Munchurl Kim.
\newblock Airsplat: Alignment and rating for robust feed-forward 3d gaussian
  splatting.
\newblock In \emph{ECCV}, 2026.

\bibitem[Chen et~al.(2025)Chen, Jiang, Jiang, Tang, Li, Liu, and
  Nie]{chen2025dashgaussian}
Youyu Chen, Junjun Jiang, Kui Jiang, Xiao Tang, Zhihao Li, Xianming Liu, and
  Yinyu Nie.
\newblock Dashgaussian: Optimizing 3d gaussian splatting in 200 seconds.
\newblock In \emph{CVPR}, 2025.

\bibitem[Dhiman et~al.(2024)Dhiman, Lu, Srinath, Arslan, Xing, Xiangli, Babu,
  and Sridhar]{dhiman2024turbo}
Ankit Dhiman, Tao Lu, R Srinath, Emre Arslan, Angela Xing, Yuanbo Xiangli,
  R~Venkatesh Babu, and Srinath Sridhar.
\newblock Turbo-gs: Accelerating 3d gaussian fitting for high-quality radiance
  fields.
\newblock \emph{arXiv preprint arXiv:2412.13547}, 2024.

\bibitem[Fan et~al.(2024)Fan, Cong, Wen, Wang, Zhang, Ding, Xu, Ivanovic,
  Pavone, Pavlakos, et~al.]{fan2024instantsplat}
Zhiwen Fan, Wenyan Cong, Kairun Wen, Kevin Wang, Jian Zhang, Xinghao Ding,
  Danfei Xu, Boris Ivanovic, Marco Pavone, Georgios Pavlakos, et~al.
\newblock Instantsplat: Sparse-view gaussian splatting in seconds.
\newblock \emph{arXiv preprint arXiv:2403.20309}, 2024.

\bibitem[Fang and Wang(2024)]{fang2024mini}
Guangchi Fang and Bing Wang.
\newblock Mini-splatting: Representing scenes with a constrained number of
  gaussians.
\newblock In \emph{ECCV}, 2024.

\bibitem[Hedman et~al.(2018)Hedman, Philip, Price, Frahm, Drettakis, and
  Brostow]{hedman2018deep}
Peter Hedman, Julien Philip, True Price, Jan-Michael Frahm, George Drettakis,
  and Gabriel Brostow.
\newblock Deep blending for free-viewpoint image-based rendering.
\newblock \emph{ACM ToG}, 2018.

\bibitem[H{\"o}llein et~al.(2025)H{\"o}llein, Bo{\v{z}}i{\v{c}}, Zollh{\"o}fer,
  and Nie{\ss}ner]{hollein20253dgs}
Lukas H{\"o}llein, Alja{\v{z}} Bo{\v{z}}i{\v{c}}, Michael Zollh{\"o}fer, and
  Matthias Nie{\ss}ner.
\newblock 3dgs-lm: Faster gaussian-splatting optimization with
  levenberg-marquardt.
\newblock In \emph{ICCV}, 2025.

\bibitem[Jena et~al.(2025)Jena, Ouasfi, Younes, and
  Boukhayma]{jena2025sparfels}
Shubhendu Jena, Amine Ouasfi, Mae Younes, and Adnane Boukhayma.
\newblock Sparfels: Fast reconstruction from sparse unposed imagery.
\newblock In \emph{ICCV}, 2025.

\bibitem[Jiang et~al.(2025)Jiang, Mao, Xu, Lu, Ren, Jin, Xu, Yu, Pang, Zhao,
  et~al.]{jiang2025anysplat}
Lihan Jiang, Yucheng Mao, Linning Xu, Tao Lu, Kerui Ren, Yichen Jin, Xudong Xu,
  Mulin Yu, Jiangmiao Pang, Feng Zhao, et~al.
\newblock Anysplat: Feed-forward 3d gaussian splatting from unconstrained
  views.
\newblock \emph{ACM ToG}, 2025.

\bibitem[Kerbl et~al.(2023)Kerbl, Kopanas, Leimk{\"u}hler, Drettakis,
  et~al.]{kerbl20233d}
Bernhard Kerbl, Georgios Kopanas, Thomas Leimk{\"u}hler, George Drettakis,
  et~al.
\newblock 3d gaussian splatting for real-time radiance field rendering.
\newblock \emph{ACM ToG}, 2023.

\bibitem[Kim et~al.(2026)Kim, Kim, Bae, Joo, and J.~Kim]{kim2026f}
Injae Kim, Chaehyeon Kim, Minseong Bae, Minseok Joo, and Hyunwoo J.~Kim.
\newblock F 4 splat: Feed-forward predictive densification for feed-forward 3d
  gaussian splatting.
\newblock In \emph{ECCV}, 2026.

\bibitem[Li et~al.(2026)Li, Gao, Wu, Chomko, Zelek, and Li]{li2026wider}
Daisy Li, Kyle Gao, Quanyun Wu, Hanna Chomko, John~S Zelek, and Jonathan Li.
\newblock When wider views fail: Stress-testing feed-forward 3d reconstruction.
\newblock \emph{arXiv preprint arXiv:2609.24839}, 2026.

\bibitem[Lin et~al.(2026)Lin, Chen, Liew, Chen, Li, Shi, Feng, and
  Kang]{lin2025depth}
Haotong Lin, Sili Chen, Junhao Liew, Donny~Y Chen, Zhenyu Li, Guang Shi, Jiashi
  Feng, and Bingyi Kang.
\newblock Depth anything 3: Recovering the visual space from any views.
\newblock In \emph{ICLR}, 2026.

\bibitem[Liu et~al.(2025)Liu, Luo, Tang, Peng, and Zhang]{liu2025vggt}
Yang Liu, Chuanchen Luo, Zimo Tang, Junran Peng, and Zhaoxiang Zhang.
\newblock Vggt-x: When vggt meets dense novel view synthesis.
\newblock \emph{arXiv preprint arXiv:2509.25191}, 2025.

\bibitem[Lyu et~al.(2026)Lyu, Tewari, Chen, Leimk{\"u}hler, and
  Theobalt]{lyu2026faster}
Linjie Lyu, Ayush Tewari, Jianchun Chen, Thomas Leimk{\"u}hler, and Christian
  Theobalt.
\newblock Faster 3d gaussian splatting convergence via structure-aware
  densification.
\newblock \emph{arXiv preprint arXiv:2604.28016}, 2026.

\bibitem[Ma et~al.(2026)Ma, Lin, Wang, Dong, Wang, Wu, and Zhang]{ma2026ba}
Zhongjie Ma, Di Lin, Xin Wang, Haotian Dong, Chong Wang, Dongdong Wu, and
  Changqing Zhang.
\newblock Ba-gs: Bayesian adaptive gaussian splatting for sfm-free 3d
  reconstruction.
\newblock In \emph{CVPR}, 2026.

\bibitem[Mallick et~al.(2024)Mallick, Goel, Kerbl, Steinberger, Carrasco, and
  De~La~Torre]{mallick2024taming}
Saswat~Subhajyoti Mallick, Rahul Goel, Bernhard Kerbl, Markus Steinberger,
  Francisco~Vicente Carrasco, and Fernando De~La~Torre.
\newblock Taming 3dgs: High-quality radiance fields with limited resources.
\newblock In \emph{SIGGRAPH}, 2024.

\bibitem[Merrell et~al.(2007)Merrell, Akbarzadeh, Wang, Mordohai, Frahm, Yang,
  Nist{\'e}r, and Pollefeys]{merrell2007real}
Paul Merrell, Amir Akbarzadeh, Liang Wang, Philippos Mordohai, Jan-Michael
  Frahm, Ruigang Yang, David Nist{\'e}r, and Marc Pollefeys.
\newblock Real-time visibility-based fusion of depth maps.
\newblock In \emph{ICCV}, 2007.

\bibitem[Pan et~al.(2024)Pan, Bar{\'a}th, Pollefeys, and
  Sch{\"o}nberger]{pan2024global}
Linfei Pan, D{\'a}niel Bar{\'a}th, Marc Pollefeys, and Johannes~L
  Sch{\"o}nberger.
\newblock Global structure-from-motion revisited.
\newblock In \emph{ECCV}, 2024.

\bibitem[Park et~al.(2025)Park, Ryu, and Kim]{park2025dropgaussian}
Hyunwoo Park, Gun Ryu, and Wonjun Kim.
\newblock Dropgaussian: Structural regularization for sparse-view gaussian
  splatting.
\newblock In \emph{CVPR}, 2025.

\bibitem[Park et~al.(2026)Park, Bui, Bello, Moon, Oh, and
  Kim]{park2025ecosplat}
Jongmin Park, Minh-Quan~Viet Bui, Juan Luis~Gonzalez Bello, Jaeho Moon, Jihyong
  Oh, and Munchurl Kim.
\newblock Ecosplat: Efficiency-controllable feed-forward 3d gaussian splatting
  from multi-view images.
\newblock In \emph{CVPR}, 2026.

\bibitem[Ren et~al.(2026)Ren, Wen, Fang, and Lu]{ren2026fastgs}
Shiwei Ren, Tianci Wen, Yongchun Fang, and Biao Lu.
\newblock Fastgs: Training 3d gaussian splatting in 100 seconds.
\newblock In \emph{CVPR}, 2026.

\bibitem[Schonberger and Frahm(2016)]{schonberger2016structure}
Johannes~L Schonberger and Jan-Michael Frahm.
\newblock Structure-from-motion revisited.
\newblock In \emph{CVPR}, 2016.

\bibitem[Sch{\"o}nberger et~al.(2016)Sch{\"o}nberger, Zheng, Frahm, and
  Pollefeys]{schonberger2016pixelwise}
Johannes~L Sch{\"o}nberger, Enliang Zheng, Jan-Michael Frahm, and Marc
  Pollefeys.
\newblock Pixelwise view selection for unstructured multi-view stereo.
\newblock In \emph{ECCV}, 2016.

\bibitem[Shade et~al.(1998)Shade, Gortler, He, and Szeliski]{shade1998layered}
Jonathan Shade, Steven Gortler, Li-wei He, and Richard Szeliski.
\newblock Layered depth images.
\newblock In \emph{SIGGRAPH}, 1998.

\bibitem[Tang et~al.(2025)Tang, Guo, Li, and Peng]{tang2025spars3r}
Yutao Tang, Yuxiang Guo, Deming Li, and Cheng Peng.
\newblock Spars3r: Semantic prior alignment and regularization for sparse 3d
  reconstruction.
\newblock In \emph{CVPR}, 2025.

\bibitem[Wang et~al.(2025)Wang, Chen, Karaev, Vedaldi, Rupprecht, and
  Novotny]{wang2025vggt}
Jianyuan Wang, Minghao Chen, Nikita Karaev, Andrea Vedaldi, Christian
  Rupprecht, and David Novotny.
\newblock Vggt: Visual geometry grounded transformer.
\newblock In \emph{CVPR}, 2025.

\bibitem[Wang et~al.(2026{\natexlab{a}})Wang, Chen, Zhang, Karaev,
  Sch{\"o}nberger, Labatut, Bojanowski, Novotny, Vedaldi, and
  Rupprecht]{wang2026vggt}
Jianyuan Wang, Minghao Chen, Shangzhan Zhang, Nikita Karaev, Johannes
  Sch{\"o}nberger, Patrick Labatut, Piotr Bojanowski, David Novotny, Andrea
  Vedaldi, and Christian Rupprecht.
\newblock {VGGT-$\Omega$}.
\newblock In \emph{CVPR}, 2026{\natexlab{a}}.

\bibitem[Wang et~al.(2026{\natexlab{b}})Wang, Zhou, Zhu, Chang, Zhou, Li, Chen,
  Pang, Shen, and He]{wang2026pi}
Yifan Wang, Jianjun Zhou, Haoyi Zhu, Wenzheng Chang, Yang Zhou, Zizun Li, Junyi
  Chen, Jiangmiao Pang, Chunhua Shen, and Tong He.
\newblock {$\pi^3$}: Permutation-equivariant visual geometry learning.
\newblock In \emph{ICLR}, 2026{\natexlab{b}}.

\bibitem[Wang et~al.(2004)Wang, Bovik, Sheikh, and Simoncelli]{wang2004image}
Zhou Wang, Alan~C Bovik, Hamid~R Sheikh, and Eero~P Simoncelli.
\newblock Image quality assessment: from error visibility to structural
  similarity.
\newblock \emph{IEEE TIP}, 2004.

\bibitem[Xu et~al.(2026)Xu, Zhan, Wei, Pfister, and Lu]{xu2026avsplat}
Muyu Xu, Fangneng Zhan, Yu Wei, Hanspeter Pfister, and Shijian Lu.
\newblock Avsplat: Dense-view feed-forward 3d gaussian splatting with
  assist-view preconditioning.
\newblock In \emph{ECCV}, 2026.

\bibitem[Ye et~al.(2026)Ye, Chen, Xu, Barath, and Pollefeys]{ye2026yonosplat}
Botao Ye, Boqi Chen, Haofei Xu, Daniel Barath, and Marc Pollefeys.
\newblock Yonosplat: You only need one model for feedforward 3d gaussian
  splatting.
\newblock In \emph{ICLR}, 2026.

\bibitem[Zhang et~al.(2024)Zhang, Li, Yu, Huang, Gu, Zheng, and
  Bai]{zhang2024cor}
Jiawei Zhang, Jiahe Li, Xiaohan Yu, Lei Huang, Lin Gu, Jin Zheng, and Xiao Bai.
\newblock Cor-gs: sparse-view 3d gaussian splatting via co-regularization.
\newblock In \emph{ECCV}, 2024.

\bibitem[Zhang et~al.(2018)Zhang, Isola, Efros, Shechtman, and
  Wang]{zhang2018unreasonable}
Richard Zhang, Phillip Isola, Alexei~A Efros, Eli Shechtman, and Oliver Wang.
\newblock The unreasonable effectiveness of deep features as a perceptual
  metric.
\newblock In \emph{CVPR}, 2018.

\bibitem[Zhao and Zelazo(2015)]{zhao2015bearing}
Shiyu Zhao and Daniel Zelazo.
\newblock Bearing rigidity and almost global bearing-only formation
  stabilization.
\newblock \emph{IEEE Trans. Autom. Control}, 2015.

\end{thebibliography}
}

\clearpage
\maketitlesupplementary

\makeatletter
\addtocontents{toc}{\protect\supp@tocstart}
\section*{Contents}
\begingroup
\let\supp@contentsline\contentsline
\def\contentsline#1#2#3#4{}
\def\supp@tocstart{\let\contentsline\supp@contentsline}
\def\numberline#1\hskip#2.{\hb@xt@\@tempdima{#1\hfil}\supp@gobbletie}
\def\supp@gobbletie{\@ifnextchar~{\@gobble}{\@ifnextchar\nobreakspace{\@gobbletwo}{}}}
\@starttoc{toc}
\endgroup
\makeatother

\renewcommand{\dbltopfraction}{0.7}
\renewcommand{\dblfloatpagefraction}{0.6}

\providecommand{\venue}[1]{{\scriptsize [#1]}}

\section{Implementation Details}
\label{sec:supp_impl}

\subsection{Camera-Perturbation-based Initialization (CPI)}
\label{sec:supp_impl_pci}

Before training, CPI refines the camera poses predicted by the 3D-VFM if the camera graph passes the bearing-rigidity test. It then removes off-surface and duplicate points from the point cloud derived from the 3D-VFM predictions (Alg.~\ref{alg:pci}).

\subsubsection{Rigidity-gated refinement}
\label{sec:supp_impl_pci_refinement}
\label{sec:supp_impl_pci_graph}
\label{sec:supp_pci_rigidity}
\label{sec:supp_pci_alignment}
\label{sec:supp_pci_depth}
Rigidity-gated refinement addresses geometric ambiguity in the camera poses predicted by the 3D-VFM~\cite{lin2025depth}. The camera pose refinement can misalign the scene geometry across views if the pairwise bearings do not constrain the relative position of every camera. We therefore apply a bearing-rigidity test~\cite{zhao2015bearing} to the camera graph to decide whether to refine the camera poses (Alg.~\ref{alg:pci}, line~\ref{line:pci_gate}).

The pairwise bearing of two cameras \(i\) and \(j\) is the unit vector
from the camera position \(o_i=-R_i^\top t_i\) to \(o_j\), where
\((R_i,t_i)\) are the world-to-camera rotation and translation of camera
\(i\). The bearing-rigidity test uses only the pairwise bearings along the
edges of the camera graph. The camera graph
has an edge \((i,j)\in\mathcal E\) for each pair of the \(V\) input cameras
whose orientations differ by less than \(30^\circ\) as in
VGGT-X~\cite{liu2025vggt}. The criterion of the bearing-rigidity test is
\(\operatorname{rank}(B)=3V-4\), where
\(B\) is the Jacobian of the pairwise bearings along the edges in
\(\mathcal E\)
with respect to the camera positions~\cite{zhao2015bearing}. This criterion
requires the pairwise bearings to constrain the relative position of every
camera.

When the camera graph passes the bearing-rigidity test, we refine the
camera poses
through the epipolar global alignment of VGGT-X~\cite{liu2025vggt}
(Alg.~\ref{alg:pci}, lines~\ref{line:pci_gate}--\ref{line:pci_refine_end}).
We match each camera pair in \(\mathcal E\) with XFeat~\citesupp{potje2024xfeat}, using up to 4,096
features. The epipolar global alignment optimizes the camera poses and a
shared focal
length for 300 Adam steps, starting from the 3D-VFM predictions.
Using the
refined cameras \(C_i^\star=(K_i^\star,R_i^\star,t_i^\star)\), we
triangulate the matches as in COLMAP~\cite{schonberger2016structure} and
update the 3D-VFM depth map of each view \(i\) to \(D_i^\star\).

\subsubsection{Camera-perturbation-based filtering}
\label{sec:supp_pci_filtering}
\label{sec:supp_pci_perturbation}
\label{sec:supp_pci_warping}
\label{sec:supp_pci_voting}

Camera-perturbation-based filtering removes off-surface point clouds by
additionally using perturbed cameras to expose depth errors
(Alg.~\ref{alg:pci}, lines~\ref{line:pci_perturb}--\ref{line:pci_vote}).
We translate
the input camera \(C_i^\star\) parallel to its image plane to obtain the
perturbed camera \(C'_i\). The translation has magnitude \(b_i\) and a
random direction \(\theta_i\sim\mathcal U[0,2\pi)\). We set \(b_i\) so that
two 3D points on a
ray of the input camera at depths \(m_i\) and \((1-\tau)m_i\) are projected
\(p_{\mathrm{tar}}\) pixels apart in the perturbed camera:
\begin{equation}
\label{eq:supp_perturbation_magnitude}
\begin{aligned}
b_i
  &= \frac{p_{\mathrm{tar}}\,m_i(1-\tau)}{\tau f_i(\theta_i)}, \\
f_i(\theta_i)
  &= \sqrt{(f_{x,i}\cos\theta_i)^2+(f_{y,i}\sin\theta_i)^2},
\end{aligned}
\end{equation}
where \(m_i\) is the median depth of \(D_i^\star\), \(\tau=0.05\) is the
relative depth tolerance, \(p_{\mathrm{tar}}=2\), and \(f_{x,i}\) and
\(f_{y,i}\) are the focal lengths of \(K_i^\star\). We forward-warp
\(D_i^\star\)~\cite{shade1998layered} to obtain the depth map
\(\widetilde D_i^\star\) of the perturbed camera.

We verify the depth of each pixel \(u\) of view \(i\) against every other
view \(j\), using both the input camera \(C_j^\star\) and the perturbed
camera \(C'_j\). We project the 3D point of \(u\) at depth
\(D_i^\star(u)\) into both cameras. For each view \(j\), we count a
deletion vote if the 3D point lies in front of the observed surface in
either camera, \textit{i.e.}, \(z<(1-\tau)d\). Here, \(z\) is the depth of
the 3D point in a camera and \(d\) is the value of the depth map of this camera,
\(D_j^\star\) or \(\widetilde D_j^\star\), at the projected location. We
count a supporting vote if the 3D point lies on the observed surface in both
cameras, \textit{i.e.}, \(\lvert z-d\rvert\leq\tau d\). We discard the pixels with more
deletion votes than supporting votes and retain the rest in \(\mathcal Q\)
(Alg.~\ref{alg:pci}, line~\ref{line:pci_vote}).

\subsubsection{Visibility-based consolidation}
\label{sec:supp_pci_consolidation}
\label{sec:supp_pci_fusion}
\label{sec:supp_pci_fillin}
\label{sec:supp_pci_sampling}

Visibility-based consolidation removes duplicate point clouds, which arise
when multiple input views observe the same surface (Alg.~\ref{alg:pci},
lines~\ref{line:pci_fusion}--\ref{line:pci_sample}). To find the duplicate
point clouds, we project the 3D point of each pixel \(u\) of view \(i\) at
depth \(D_i^\star(u)\) into every other view \(j\). The point cloud of view
\(j\) contains a duplicate 3D point if the 3D point of \(u\) lies on the
surface observed in view \(j\), \textit{i.e.},
\(\lvert z-d\rvert<\tau d\). Here, \(z\) is the depth of the 3D point of
\(u\) in view \(j\) and \(d\) is the value of \(D_j^\star\) at the
projected location. Visibility-based fusion~\cite{merrell2007real} merges
the 3D point of \(u\) and the duplicate 3D points into a single 3D point,
which lies on the ray of \(u\) at their mean depth in view \(i\). We count
a conflict if the merged 3D point occludes the surface observed in another
view, or if a 3D point of another view occludes the merged 3D point in
view \(i\). We keep the merged 3D point only if the 3D points before
merging include 3D points of at least two other views and outnumber the
conflicts~\cite{merrell2007real}.

Where no 3D point remains after merging, we use the unchanged 3D points of
the pixels in \(\mathcal Q\) (Alg.~\ref{alg:pci},
line~\ref{line:pci_vote}). To limit the number of initial Gaussians, we uniformly subsample
the resulting point cloud to a size of at most 500,000, as in
VGGT-X~\cite{liu2025vggt} (Alg.~\ref{alg:pci}, line~\ref{line:pci_sample}).
The sampled point cloud \(S\) initializes the Gaussians with the pixel
colors \(\mathbf c_S\) (Alg.~\ref{alg:pci}, line~\ref{line:pci_return}).

\subsection{Camera-Perturbation-based Adaptive density control (CPA)}
\label{sec:supp_cpa}

During training, CPA removes the off-surface Gaussians that remain after
CPI and limits the redundant child Gaussians that per-Gaussian
densification adds.

\subsubsection{Contribution-based pruning}
\label{sec:supp_cpa_pruning}
\label{sec:supp_cpa_contribution}
\label{sec:supp_cpa_pruning_rule}

Contribution-based pruning removes the Gaussians with a low total rendering
contribution \(A_j\) across the input and perturbed views
(Eq.~\eqref{eq:cpa_contribution}). The perturbed camera \(C'_k\) of each
input view \(k\) is obtained from the camera \(C_k^\star\) of this view as
in CPI (Sec.~\ref{sec:supp_pci_filtering}), with the same magnitude \(b_k\)
(Eq.~\eqref{eq:supp_perturbation_magnitude}) and direction \(\theta_k\). We remove the Gaussians in ascending order of
\(A_j\) until the removed Gaussians account for a fraction \(\rho=0.1\) of
the total rendering contribution.

\subsubsection{Group-wise densification}
\label{sec:supp_cpa_densification}
\label{sec:supp_cpa_grouping}
\label{sec:supp_cpa_budget}
\label{sec:supp_cpa_priority}

The split rule of SAD-GS~\cite{lyu2026faster} estimates the number \(n\) of
child Gaussians that a Gaussian needs to represent the local image detail:
\begin{equation}
\label{eq:supp_split_rule}
n=\prod_{a=1}^{3}\left\lceil\sqrt{\max(1,\eta_a)}\right\rceil,
\end{equation}
where \(\eta_a\) is the ratio of the length of principal axis \(a\) of the
Gaussian in the image to the wavelength of the local image detail.
Per-Gaussian densification applies the split rule to each Gaussian
separately, so it produces redundant child Gaussians that represent the
same image detail. Group-wise densification therefore groups the Gaussians
that represent the same image detail into a query \(q\) and limits the
number of Gaussians in each query to a budget \(N_q\).

We group the Gaussians that project to the same region, since these
Gaussians represent the same image detail. Specifically, in view \(k\), a
query \(q\) is the group of Gaussians that project to the same tile of
\(16\times16\) pixels~\cite{kerbl20233d} and whose image-plane
displacements \(\delta\) between view \(k\) and the perturbed view lie in
the same one-pixel interval, \textit{i.e.}, have the same
\(\lfloor\delta\rfloor\). Here, view \(k\) is the input view to which a
Gaussian contributes most. The image-plane displacement
\(\delta=b_k f_k(\theta_k)/z\) is the distance in pixels between the
projections of the Gaussian center in \(C_k^\star\) and in the perturbed
camera \(C'_k\), where \(z\) is the depth of the Gaussian center in
\(C_k^\star\) and \(f_k(\theta_k)\) is the focal length along \(\theta_k\)
(Eq.~\eqref{eq:supp_perturbation_magnitude}).

Group-wise densification applies the split rule to each query as a whole
instead of to each Gaussian separately. Specifically, the budget \(N_q\) is
the number \(n\) of child Gaussians that the split rule estimates for the
representative Gaussian \(g_q\) of the query
(Eq.~\eqref{eq:supp_split_rule}). Here, \(g_q\) is a single
Gaussian that approximates the region occupied by all Gaussians in the
query. The center and covariance of \(g_q\) are the mean and covariance of
the Gaussians in the query, weighted by the rendering contributions
\(A_j\). Within each query, we first densify the Gaussians that are largest
relative to the local image detail, \textit{i.e.}, in descending order of
the priority \(\prod_{a=1}^{3}\max(1,\eta_a)\). We stop before the number
of Gaussians in the query exceeds \(N_q\).

\begin{table}[!t]
    \centering
    \caption{\textbf{Datasets and view settings.}
     Evaluation resolution, total number of images, and number of test views for each scene.}
    \label{tab:supp_datasets}

    \footnotesize
    \setlength{\tabcolsep}{10pt}
    \begin{tabular}{@{}lccc@{}}
        \toprule
        Scene & Resolution (px) & Images & Test views \\
        \midrule
        \multicolumn{4}{c}{\textbf{Mip-NeRF~360}~\cite{barron2022mip}} \\
        \midrule
        \textit{bicycle}  & $618\times411$ & 194 & 25 \\
        \textit{bonsai}   & $390\times260$ & 292 & 37 \\
        \textit{counter}  & $389\times260$ & 240 & 30 \\
        \textit{flowers}  & $628\times414$ & 173 & 22 \\
        \textit{garden}   & $648\times420$ & 185 & 24 \\
        \textit{kitchen}  & $389\times260$ & 279 & 35 \\
        \textit{room}     & $389\times259$ & 311 & 39 \\
        \textit{stump}    & $622\times413$ & 125 & 16 \\
        \textit{treehill} & $634\times416$ & 141 & 18 \\
        \midrule
        \multicolumn{4}{c}{\textbf{Deep Blending}~\cite{hedman2018deep}} \\
        \midrule
        \textit{drjohnson} & $1332\times876$ & 263 & 33 \\
        \textit{playroom}  & $1264\times832$ & 225 & 29 \\
        \bottomrule
    \end{tabular}
\end{table}

\begin{table*}[tp]
    \centering
    \caption{\textbf{Settings of the comparison methods.}
    The rows (a)--(g) are those of Tab.~\ref{tab:view_count}.
    $^{\star}$ and $^{\diamond}$ denote COLMAP~\cite{schonberger2016structure,schonberger2016pixelwise} and GLOMAP~\cite{pan2024global}.
    ``Official'' denotes the code released by the authors of each method, run with the settings of the release.}
    \label{tab:supp_comparison_settings}

    \footnotesize
    \setlength{\tabcolsep}{4pt}
    \begin{tabular}{@{}cllll@{}}
        \toprule
        & Method & Camera poses and point clouds & 3DGS training & Change from the official setting \\
        \midrule
        (a) & FastGS$^{\star}$ \venue{CVPR'26}~\cite{ren2026fastgs} & COLMAP~\cite{schonberger2016structure,schonberger2016pixelwise} on the input views & official & -- \\
        (b) & FastGS$^{\diamond}$ \venue{CVPR'26}~\cite{ren2026fastgs} & GLOMAP~\cite{pan2024global} on the input views & official & -- \\
        (c) & SAD-GS$^{\star}$ \venue{SIGGRAPH'26}~\cite{lyu2026faster} & COLMAP~\cite{schonberger2016structure,schonberger2016pixelwise} on the input views & official & -- \\
        (d) & SAD-GS$^{\diamond}$ \venue{SIGGRAPH'26}~\cite{lyu2026faster} & GLOMAP~\cite{pan2024global} on the input views & official & -- \\
        \midrule
        (e) & InstantSplat \venue{arXiv'24}~\cite{fan2024instantsplat} & official & official & -- \\
        (f) & VGGT-X \venue{arXiv'25}~\cite{liu2025vggt} & official & official & maximum number of Gaussians set to that of our \ours{} \\
        (g) & AirSplat \venue{ECCV'26}~\cite{bui2026airsplat} & feed-forward prediction & none & -- \\
        \bottomrule
    \end{tabular}
\end{table*}

\subsection{Training and Evaluation Settings}
\label{sec:supp_training}
\label{sec:supp_training_config}
\label{sec:supp_hyperparameters}
\label{sec:supp_eval_protocol}
We initialize from the predictions of DA3-GIANT-1.1~\cite{lin2025depth} and jointly optimize the Gaussians and input camera poses for 10,000 iterations. We parameterize each pose correction with a quaternion and a translation vector and compose it with the initial world-to-camera transform.

We use the same settings across all input view counts. CPI uses a target parallax of \(p_{\mathrm{tar}}=2\) pixels on the depth grid (Sec.~\ref{sec:supp_pci_perturbation}) and a relative depth tolerance of \(\tau=0.05\), as used to assess depth agreement in visibility-based fusion~\cite{merrell2007real}. Following the 500-iteration densification interval of SAD-GS~\cite{lyu2026faster}, CPA applies contribution-based pruning and group-wise densification at nine steps from iteration 1,000 through 5,000. At each step, the target cumulative rendering contribution of the Gaussians selected for pruning is a fraction \(\rho=0.1\) of the total rendering contribution. Fig.~\ref{fig:supp_pruning_fraction} evaluates the effect of varying \(\rho\).

Tab.~\ref{tab:supp_datasets} lists the scenes and evaluation resolutions. We use all nine Mip-NeRF~360 scenes~\cite{barron2022mip} at \(8\times\) downsampling and both Deep Blending scenes~\cite{hedman2018deep} at their native resolutions. Following DropGaussian~\cite{park2025dropgaussian}, we select evenly spaced training indices and round each to the nearest integer. When fewer training images are available than the specified number of input views, we use all available training images.

We use the selected training images to reconstruct each scene. Tab.~\ref{tab:supp_comparison_settings} lists the initialization and training settings for the methods compared in Tab.~\ref{tab:view_count}. For evaluation, we obtain reference test poses from a separate SfM registration that uses all views. We then map these poses into the reconstruction coordinate system using a similarity transform fitted to the corresponding training cameras~\cite{tang2025spars3r}. For SfM-free methods, we further refine each mapped pose using its corresponding test image while keeping the Gaussians fixed~\cite{fan2024instantsplat}. We evaluate PSNR, SSIM~\cite{wang2004image}, and VGG-LPIPS~\cite{zhang2018unreasonable} on the full images.

\section{Additional Qualitative Comparisons}
\label{sec:supp_demo}
\label{sec:supp_additional_qualitative}
\label{sec:supp_qualitative}

\begin{table*}[tp]
    \centering
    \caption{\textbf{Direct combination of 3D-VFM initialization and 3DGS training acceleration on Mip-NeRF~360.}
    Row (a) initializes one Gaussian from every pixel-wise prediction of the 3D-VFM and trains the Gaussians with SAD-GS~\cite{lyu2026faster} in the official setting.
    Row (b) randomly subsamples the initialization of (a) to 0.5M Gaussians.
    Both rows refine the camera poses of the 3D-VFM during 3DGS training, as our \ours{} does.
    We report PSNR$\uparrow$/SSIM$\uparrow$/LPIPS$\downarrow$, the number of initial Gaussians in millions (\#Init.~G$\downarrow$), and the total reconstruction time in minutes (Time$\downarrow$).}
    \label{tab:supp_vfm_acceleration}

    \footnotesize
    \setlength{\tabcolsep}{3pt}
    \resizebox{\textwidth}{!}{%
    \begin{tabular}{@{}cl*{6}{c}@{}}
        \toprule
        & & \multicolumn{3}{c}{(i) 24 views}
        & \multicolumn{3}{c}{(ii) 192 views} \\
        \cmidrule(lr){3-5}
        \cmidrule(lr){6-8}
        & Variant
        & PSNR$\uparrow$/SSIM$\uparrow$/LPIPS$\downarrow$ & \#Init.~G$\downarrow$ (M) & Time$\downarrow$ (min)
        & PSNR$\uparrow$/SSIM$\uparrow$/LPIPS$\downarrow$ & \#Init.~G$\downarrow$ (M) & Time$\downarrow$ (min) \\
        \midrule
        (a) & 3D-VFM init.~\cite{lin2025depth} + SAD-GS \venue{SIGGRAPH'26}~\cite{lyu2026faster}
        & 19.46/0.505/0.352 & 4.05 & 1.1 & OOM & 27.74 & -- \\
        (b) & (a) + random subsampling
        & 20.67/0.566/0.301 & 0.50 & 0.7 & 25.82/0.770/0.183 & 0.50 & 1.1 \\
        \midrule
        \rowcolor{gray!15}
        (c) & \textbf{\ours{} (Ours)}
        & 21.46/0.620/0.269 & 0.50 & 1.9 & 27.51/0.843/0.136 & 0.50 & 3.0 \\
        \bottomrule
    \end{tabular}
    }
\end{table*}

\begin{figure*}[tp]
    \centering
    \includegraphics[width=\textwidth]{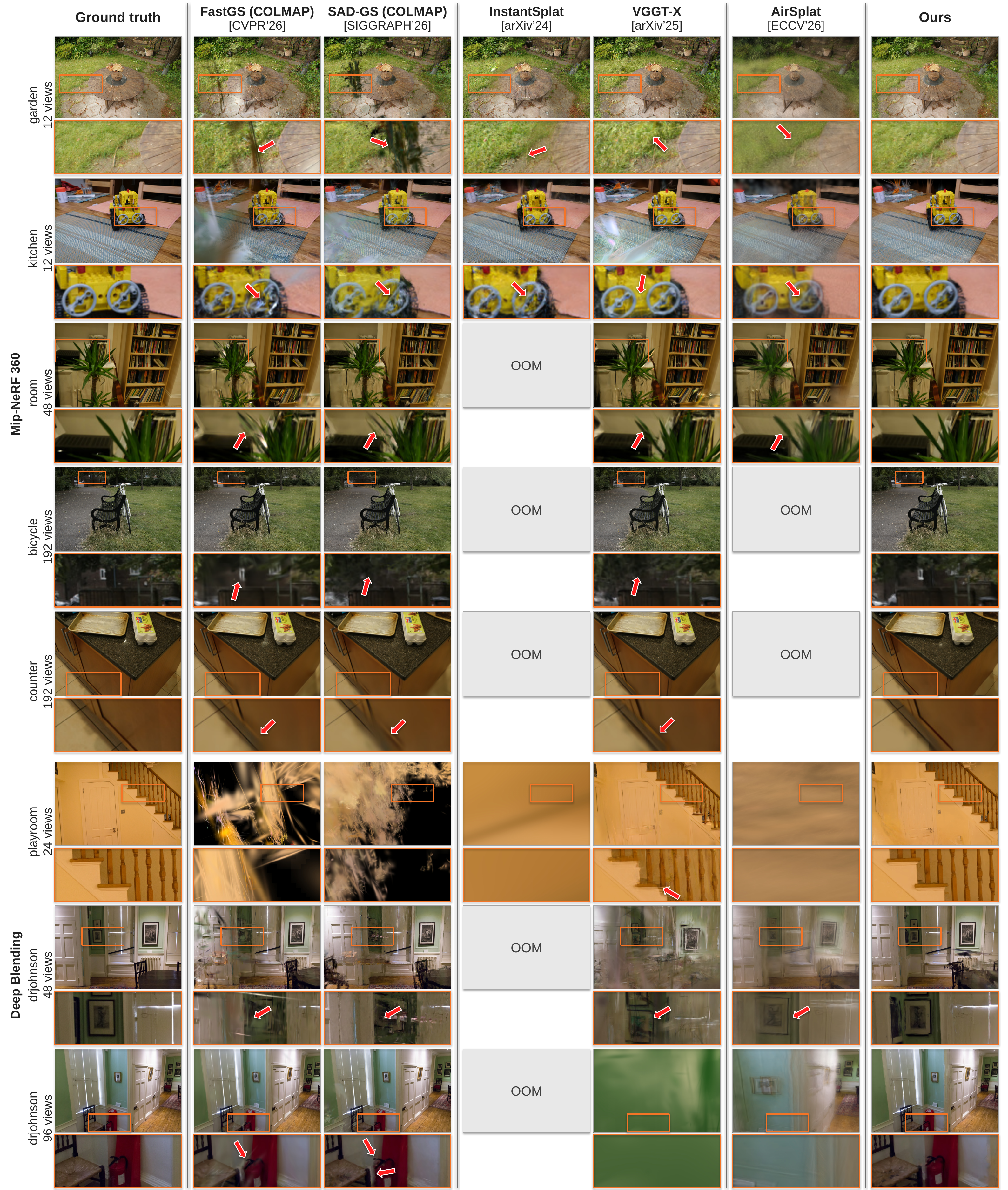}
    \caption{\textbf{Additional qualitative comparison on Mip-NeRF~360 and Deep Blending.}
    Vertical lines separate the methods for 3DGS training acceleration, 3D-VFM initialization, and feed-forward 3DGS. The enlarged view of the marked region is shown below each rendering, and the red arrows mark the artifacts of the compared methods. OOM denotes an out-of-memory error.}
    \label{fig:supp_qualitative_1}
\end{figure*}

Fig.~\ref{fig:supp_qualitative_1} extends the qualitative comparison of
Fig.~\ref{fig:qualitative} to further scenes and test views of
Mip-NeRF~360 and Deep Blending, from 12 to 192 input views. In the marked regions, the compared methods show floaters, missing structure, or blur, where our \ours{} preserves the structure. Tab.~\ref{tab:view_count} reports
the averages over all test views.

\section{Additional Results}
\label{sec:supp_results}

\subsection{Direct Combination of 3D-VFM Initialization and 3DGS Training Acceleration}
\label{sec:supp_vfm_acceleration}

Directly combining DA3 initialization with SAD-GS leads to an
out-of-memory error in the 192-view setting
(Tab.~\ref{tab:supp_vfm_acceleration}(a)), showing that dense
initialization can prevent the accelerated training pipeline from
completing. Random subsampling makes training feasible
(Tab.~\ref{tab:supp_vfm_acceleration}(b)), but controls only the
number of initial Gaussians without addressing geometric errors or
redundant observations. With the same initial Gaussian count,
\ours{} achieves better reconstruction quality at the cost of
additional reconstruction time
(Tab.~\ref{tab:supp_vfm_acceleration}(c)). This comparison shows
that matching the initialization size alone does not close the
quality gap between the direct combination and our full pipeline.

\subsection{Computational Efficiency}
\label{sec:supp_efficiency}
\label{sec:supp_initial_gaussians}
\label{sec:supp_runtime_memory}

Tab.~\ref{tab:supp_efficiency} reports runtime by stage, peak GPU memory, and the final Gaussian counts after training. Stage~II (3DGS training with CPA) accounts for the largest share of runtime at every view count on both datasets. Increasing the number of input views from 12 to 192 raises total runtime by 50\% on Mip-NeRF~360 and 34\% on Deep Blending. Thus, a 16-fold increase in input views leads to at most a 1.5-fold increase in runtime, while peak GPU memory grows by a factor of 2.6 on both datasets.

Tab.~\ref{tab:supp_point_cloud_size} shows how CPI controls the size of the point cloud before 3DGS training. At 192 views, visibility-based consolidation reduces the number of points retained after camera-perturbation-based filtering by a further 78\% on Mip-NeRF~360 and 65\% on Deep Blending. This indicates that many observations remain redundant even after depth-inconsistent points are removed. By merging these observations before uniform subsampling, CPI incorporates evidence from additional views while keeping the initialization fixed at 0.50M Gaussians across all evaluated view counts.

\begin{table}[!t]
    \centering
    \caption{\textbf{Computational efficiency of our \ours{} with 12 to 192 input views.}
    We report the time in seconds (Time$\downarrow$) of the 3D-VFM inference, Stage~I (CPI), and Stage~II (CPA) together with their total, the number of Gaussians after 3DGS training in millions (\#G$\downarrow$), and the peak GPU memory in GB (Mem.$\downarrow$).
    Each time counts computation only and excludes file loading.
    All values are measured on a single NVIDIA RTX 5090 GPU.}
    \label{tab:supp_efficiency}

    \footnotesize
    \setlength{\tabcolsep}{5pt}
    \begin{tabular}{@{}c*{6}{c}@{}}
        \toprule
        & \multicolumn{4}{c}{Time$\downarrow$ (s)} & & \\
        \cmidrule(lr){2-5}
        Views
        & 3D-VFM & CPI & CPA & Total
        & \#G$\downarrow$ (M) & Mem.$\downarrow$ (GB) \\
        \midrule
        \multicolumn{7}{c}{\textbf{Mip-NeRF~360}} \\
        \midrule
        12  & 0.9 & 0.5 & 108.4 & 109.8 & 0.74 & 10.2 \\
        24  & 1.5 & 0.5 & 113.3 & 115.3 & 1.01 & 11.3 \\
        48  & 3.3 & 2.0 & 118.4 & 123.7 & 1.26 & 13.5 \\
        96  & 8.0 & 4.8 & 123.2 & 136.0 & 1.25 & 17.8 \\
        192 & 17.0 & 20.9 & 127.1 & 165.0 & 1.21 & 26.8 \\
        \midrule
        \multicolumn{7}{c}{\textbf{Deep Blending}} \\
        \midrule
        12  & 0.7 & 0.5 & 151.0 & 152.2 & 0.77 & 10.2 \\
        24  & 0.9 & 0.5 & 163.0 & 164.4 & 1.11 & 11.3 \\
        48  & 2.2 & 0.7 & 166.8 & 169.7 & 1.55 & 13.5 \\
        96  & 5.5 & 1.1 & 187.1 & 193.7 & 1.76 & 17.8 \\
        192 & 17.3 & 2.4 & 184.1 & 203.8 & 1.78 & 26.8 \\
        \bottomrule
    \end{tabular}
\end{table}

\begin{table}[!t]
    \centering
    \caption{\textbf{Point cloud size after each step of our CPI with 12 to 192 input views.}
    We report the size of the point cloud in millions (Point cloud size$\downarrow$) after back-projecting every pixel of the 3D-VFM depth maps (3D-VFM), after camera-perturbation-based filtering (Filtering), after the fusion and gap filling of visibility-based consolidation (Consolidation), and after uniform subsampling.
    The size of the point cloud after uniform subsampling is the number of initial Gaussians (\#Init.~G$\downarrow$).}
    \label{tab:supp_point_cloud_size}

    \footnotesize
    \setlength{\tabcolsep}{8pt}
    \begin{tabular}{@{}c*{4}{c}@{}}
        \toprule
        & \multicolumn{4}{c}{Point cloud size$\downarrow$ (M)} \\
        \cmidrule(lr){2-5}
        Views
        & 3D-VFM & Filtering & Consolidation & \#Init.~G$\downarrow$ \\
        \midrule
        \multicolumn{5}{c}{\textbf{Mip-NeRF~360}} \\
        \midrule
        12  & 2.02 & 1.66 & 1.10 & 0.50 \\
        24  & 4.05 & 3.37 & 1.63 & 0.50 \\
        48  & 8.09 & 6.40 & 2.33 & 0.50 \\
        96  & 16.18 & 12.57 & 3.42 & 0.50 \\
        192 & 27.74 & 21.73 & 4.68 & 0.50 \\
        \midrule
        \multicolumn{5}{c}{\textbf{Deep Blending}} \\
        \midrule
        12  & 2.03 & 1.82 & 1.64 & 0.50 \\
        24  & 4.06 & 3.69 & 2.90 & 0.50 \\
        48  & 8.13 & 7.35 & 4.39 & 0.50 \\
        96  & 16.26 & 14.40 & 6.61 & 0.50 \\
        192 & 32.51 & 28.38 & 9.82 & 0.50 \\
        \bottomrule
    \end{tabular}
\end{table}

\subsection{SfM Registration across View Counts}
\label{sec:supp_sfm_registration}

Tab.~\ref{tab:supp_sfm_registration} reports registration coverage and
SfM preprocessing time for the input views used in Tab.~\ref{tab:view_count}.
With 12 views of \textit{bicycle}, COLMAP registers 3 of 12 cameras and
GLOMAP registers 6 of 12. On \textit{room} at 24 views, GLOMAP registers
23 cameras, yet the mean rotation error reaches the \(20^\circ\)
threshold for a wrong reconstruction. Thus, registration coverage alone
does not establish pose accuracy. At the 192-view setting, both methods
register every available input camera on both datasets. On Mip-NeRF~360,
mean preprocessing time rises from 0.5 to 16.7 minutes for COLMAP and
from 0.4 to 7.1 minutes for GLOMAP between the 12- and 192-view settings.

\begin{table*}[tp]
    \centering
    \caption{\textbf{SfM registration with COLMAP~\cite{schonberger2016structure,schonberger2016pixelwise} and GLOMAP~\cite{pan2024global} for 12 to 192 input views.}
    Entries show registered cameras / input views ($\uparrow$).
    A dash indicates a tie in camera count between the two largest reconstructions, so neither is used.
    $^{\dagger}$ marks a wrong reconstruction (mean rotation error of registered cameras $\geq 20^\circ$).
    In Tab.~\ref{tab:view_count}, $^{\ddagger}$ marks missing scores for both FastGS and SAD-GS, and $^{\S}$ for FastGS only.
    SfM time ($\downarrow$) is the mean preprocessing time in minutes over scenes with a reconstruction.}
    \label{tab:supp_sfm_registration}

    \footnotesize
    \setlength{\tabcolsep}{4pt}
    \begin{tabular}{@{}l*{10}{c}@{}}
        \toprule
        & \multicolumn{2}{c}{(i) 12 views} & \multicolumn{2}{c}{(ii) 24 views} & \multicolumn{2}{c}{(iii) 48 views}
        & \multicolumn{2}{c}{(iv) 96 views} & \multicolumn{2}{c}{(v) 192 views} \\
        \cmidrule(lr){2-3} \cmidrule(lr){4-5} \cmidrule(lr){6-7} \cmidrule(lr){8-9} \cmidrule(lr){10-11}
        Registered$\uparrow$ / input views & COLMAP & GLOMAP & COLMAP & GLOMAP & COLMAP & GLOMAP & COLMAP & GLOMAP & COLMAP & GLOMAP \\
        \midrule
        \multicolumn{11}{c}{\textbf{Mip-NeRF~360}~\cite{barron2022mip}} \\
        \midrule
        \textit{bicycle} & 3/12 & 6/12$^{\ddagger}$ & 23/24 & 16/24 & 48/48 & 48/48 & 96/96 & 96/96 & 169/169 & 169/169 \\
        \textit{bonsai} & 10/12$^{\dagger}$ & 12/12 & 24/24 & 24/24 & 48/48 & 48/48 & 96/96 & 96/96 & 192/192 & 192/192 \\
        \textit{counter} & 12/12 & 12/12 & 24/24 & 24/24 & 48/48 & 48/48 & 96/96 & 96/96 & 192/192 & 192/192 \\
        \textit{flowers} & 5/12 & 6/12$^{\ddagger}$ & 12/24 & 18/24$^{\S}$ & 48/48 & 46/48 & 96/96 & 96/96 & 151/151 & 151/151 \\
        \textit{garden} & 12/12 & 12/12 & 24/24 & 24/24 & 48/48 & 48/48 & 96/96 & 96/96 & 161/161 & 161/161 \\
        \textit{kitchen} & 12/12 & 12/12 & 24/24 & 23/24 & 48/48 & 48/48 & 96/96 & 96/96 & 192/192 & 192/192 \\
        \textit{room} & 6/12 & -- & 24/24 & 23/24$^{\dagger}$ & 48/48 & 48/48 & 96/96 & 96/96 & 192/192 & 192/192 \\
        \textit{stump} & 4/12 & 7/12 & 21/24 & 21/24 & 48/48 & 48/48 & 96/96 & 96/96 & 109/109 & 109/109 \\
        \textit{treehill} & 7/12 & 6/12 & 24/24 & 24/24 & 48/48 & 48/48 & 96/96 & 96/96 & 123/123 & 123/123 \\
        \cmidrule{1-11}
        SfM time$\downarrow$ (min) & 0.5 & 0.4 & 1.1 & 0.7 & 2.3 & 1.4 & 6.7 & 3.2 & 16.7 & 7.1 \\
        \midrule
        \multicolumn{11}{c}{\textbf{Deep Blending}~\cite{hedman2018deep}} \\
        \midrule
        \textit{drjohnson} & 5/12$^{\dagger\S}$ & -- & 16/24 & 22/24$^{\dagger}$ & 45/48$^{\dagger}$ & 48/48$^{\ddagger}$ & 95/96 & 95/96 & 192/192 & 192/192 \\
        \textit{playroom} & 6/12 & 9/12$^{\ddagger}$ & 10/24 & 24/24$^{\ddagger}$ & 48/48 & 48/48 & 96/96 & 95/96 & 192/192 & 192/192 \\
        \cmidrule{1-11}
        SfM time$\downarrow$ (min) & 0.1 & 0.1 & 0.2 & 0.1 & 0.6 & 0.2 & 1.5 & 0.7 & 5.2 & 1.9 \\
        \bottomrule
    \end{tabular}
\end{table*}

\section{Additional Analysis}
\label{sec:supp_analysis}

\subsection{Pruning with and without Perturbed Views}
\label{sec:supp_pruning_comparison}

Tab.~\ref{tab:supp_pruning_comparison} compares contribution-based pruning with and without perturbed views. Row~(a) computes the contribution score \(A_j\) using only the input views \(C_k^\star\). Row~(b) uses CPA, which computes \(A_j\) over both the input views \(C_k^\star\) and their perturbed views \(C'_k\) (Eq.~\eqref{eq:cpa_contribution}). At each of the nine pruning steps, both variants start from the same Gaussians and use the same cumulative contribution fraction \(\rho=0.1\). We use test views only for evaluation and average the results over the nine steps.

With sparse input views, a Gaussian can contribute little at the observed camera positions yet become visible when the camera moves slightly. Scoring it from a nearby perturbed view gives CPA an opportunity to retain it. The upper block of Tab.~\ref{tab:supp_pruning_comparison} shows that the Gaussians removed by CPA in row~(b) contribute less to held-out views than those removed by input-only pruning in row~(a). The lower block shows the corresponding benefit in rendering quality. CPA reduces the immediate PSNR drop at all five view counts, with a 23.3\% reduction at 12 views and a 1.7\% reduction at 192 views. This smaller gain with dense inputs is consistent with the broader range of viewpoints already represented in the input-only score. Fig.~\ref{fig:supp_pruning_fraction} compares the post-pruning PSNR gain of row~(b) over row~(a) across fractions \(\rho\).

\begin{table*}[tp]
    \centering
    \caption{\textbf{Contribution-based pruning with and without perturbed views on Mip-NeRF~360.}
    Row~(a) computes \(A_j\) (Eq.~\eqref{eq:cpa_contribution}) over input views only, while row~(b) also includes perturbed views (CPA).
    Both variants prune from the same Gaussians at \(\rho=0.1\) at each of nine steps. Results are averaged over these steps.
    The upper block shows the percentage of test-view rendering contribution removed. The lower block shows test-view quality immediately before and after pruning.
    Relative reductions compare row~(b) with row~(a) in terms of removed contribution and PSNR drop, respectively.
     Test views are used only for evaluation.}
    \label{tab:supp_pruning_comparison}

    \footnotesize
    \setlength{\tabcolsep}{6pt}
    \begin{tabular}{@{}cl*{5}{c}@{}}
        \toprule
        & Score & (i) 12 views & (ii) 24 views & (iii) 48 views & (iv) 96 views & (v) 192 views \\
        \midrule
        \multicolumn{7}{c}{\textbf{Removed test contribution$\downarrow$ (\%)}} \\
        \midrule
        (a) & Input views only
        & 9.98 & 9.00 & 8.50 & 8.45 & 8.47 \\
        \rowcolor{gray!15}
        (b) & \textbf{Input and perturbed views (Ours)}
        & \textbf{9.29} & \textbf{8.57} & \textbf{8.24} & \textbf{8.27} & \textbf{8.33} \\
        & Relative reduction$\uparrow$ (\%)
        & 6.9 & 4.8 & 3.1 & 2.1 & 1.7 \\
        \midrule
        \multicolumn{7}{c}{\textbf{PSNR$\uparrow$/SSIM$\uparrow$/LPIPS$\downarrow$ right after pruning}} \\
        \midrule
        & Before pruning
        & 18.05/0.460/0.386 & 20.88/0.587/0.302 & 23.05/0.658/0.258 & 23.95/0.700/0.239 & 24.06/0.708/0.246 \\
        (a) & Input views only
        & 17.75/0.448/0.404 & 20.42/0.571/0.323 & 22.40/0.638/0.282 & 23.31/0.681/0.263 & 23.47/0.688/0.269 \\
        \rowcolor{gray!15}
        (b) & \textbf{Input and perturbed views (Ours)}
        & \textbf{17.82}/\textbf{0.449}/0.404 & \textbf{20.46}/0.571/0.323 & \textbf{22.42}/\textbf{0.639}/0.282 & \textbf{23.33}/0.681/0.263 & \textbf{23.48}/0.688/0.269 \\
        & Reduction of PSNR drop$\uparrow$ (\%)
        & 23.3 & 8.7 & 3.1 & 3.1 & 1.7 \\
        \bottomrule
    \end{tabular}
\end{table*}

\begin{figure}[t]
    \centering
    \includegraphics[width=\linewidth]{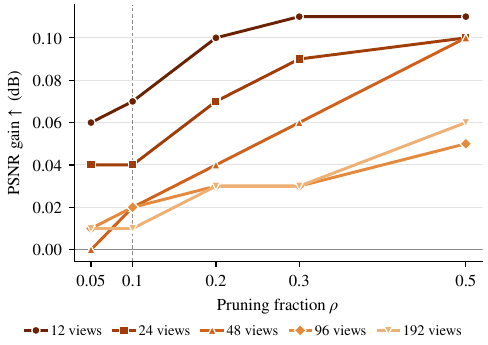}
    \caption{\textbf{Contribution-based pruning with and without perturbed views for different pruning fractions \(\rho\) on Mip-NeRF~360.}
    For each \(\rho\) and each number of input views, we plot the PSNR of row (b) minus the PSNR of row (a) of Tab.~\ref{tab:supp_pruning_comparison} on the test views right after pruning in dB, averaged over the nine pruning steps (PSNR gain$\uparrow$).
    The vertical line marks the default \(\rho=0.1\) of our CPA.}
    \label{fig:supp_pruning_fraction}
\end{figure}

\subsection{Hyperparameter Sensitivity}
\label{sec:supp_perturbation_sensitivity}
\label{sec:supp_sensitivity}
\label{sec:supp_filter_consolidation}
\label{sec:supp_pruning_budget}

Tab.~\ref{tab:supp_sensitivity} examines the effect of the target parallax \(p_{\mathrm{tar}}\) (denoted by \(p\) in the table). In CPI, the perturbed view can reveal an off-surface point hidden in an input view, allowing cross-view depth voting to flag it for removal. In CPA, Gaussians assigned to the same input view and tile share a densification budget if their projected displacements fall in the same bin. A larger target parallax makes depth differences easier to distinguish within a tile. The final Gaussian count increases slightly, while reconstruction quality remains robust across the tested range. We use \(p_{\mathrm{tar}}=2\) pixels across all view counts.

\begin{table}[H]
    \centering
    \caption{\textbf{Sensitivity to the parallax \(p\) of the camera perturbation on Mip-NeRF~360.}
    The parallax \(p\) sets the translation of each perturbed camera so that a depth 5\% nearer than the median depth of the input view shifts by \(p\) pixels relative to the median depth; our CPI and our CPA use the same perturbed cameras.
    The \textcolor{gray}{gray} row is the default setting.
    We report PSNR$\uparrow$/SSIM$\uparrow$/LPIPS$\downarrow$ and the number of Gaussians in millions (\#G$\downarrow$).}
    \label{tab:supp_sensitivity}

    \fontsize{6.5}{7.5}\selectfont
    \setlength{\tabcolsep}{1.5pt}
    \renewcommand{\arraystretch}{0.9}
    \resizebox{\columnwidth}{!}{%
    \begin{tabular}{@{}c*{4}{c}@{}}
        \toprule
        & \multicolumn{2}{c}{(i) 24 views}
        & \multicolumn{2}{c}{(ii) 192 views} \\
        \cmidrule(lr){2-3}
        \cmidrule(lr){4-5}
        \(p\) (px)
        & PSNR$\uparrow$/SSIM$\uparrow$/LPIPS$\downarrow$ & \#G$\downarrow$ (M)
        & PSNR$\uparrow$/SSIM$\uparrow$/LPIPS$\downarrow$ & \#G$\downarrow$ (M) \\
        \midrule
        1
        & 21.51/0.622/0.269 & 1.00 & 27.52/0.845/0.137 & 1.17 \\
        \rowcolor{gray!15}
        2
        & 21.46/0.620/0.269 & 1.01 & 27.51/0.843/0.136 & 1.21 \\
        4
        & 21.45/0.620/0.269 & 1.04 & 27.56/0.846/0.133 & 1.25 \\
        \bottomrule
    \end{tabular}%
    }
\end{table}

\section{Limitations}
\label{sec:supp_limitations}

With dense input views, our \ours{} does not surpass the oracles that start
from the poses and point clouds of
COLMAP~\cite{schonberger2016structure,schonberger2016pixelwise} or
GLOMAP~\cite{pan2024global} in novel view quality, but it maintains a short
total reconstruction time as the number of input views increases
(Tab.~\ref{tab:view_count}). These oracles depend on SfM, which does not
register every input camera with sparse input views
(Tab.~\ref{tab:supp_sfm_registration}).

\par
\makeatletter
\AddToHookNext{env/thebibliography/before}{\let\@auxout\@mainaux}
\makeatother

{
    \small
    \renewcommand{\bibnumfmt}[1]{[S#1]}
    \bibliographystyleS{arXiv/ieeenat_fullname}
    \bibliographyS{arXiv/supp}
}

\end{document}